\documentclass[10pt,twocolumn]{article}

\usepackage{cvpr}      % To produce the REVIEW version
\usepackage{algorithm}
\usepackage{xcolor}
\usepackage{array}
\usepackage{algpseudocode}
\usepackage{multirow}
\usepackage{multicol}
\usepackage{lipsum}
\floatname{algorithm}{Algorithm}

\usepackage{colortbl}

\newcommand{\et}{\textit{et.al.}}
\newcommand{\SEG}{\texttt{<SEG>}\ }

\newcommand{\TAK}{\texttt{<TAK>}\ }
\newcommand{\pub}[1]{\color{gray}{\tiny{[{#1}]}}}
\newcommand{\best}[1]{\cellcolor{LightBlue1}\textbf{#1}}
\newcommand{\second}[1]{\cellcolor{Bisque1}\underline{#1}}

\newlength\savewidth\newcommand\shline{\noalign{\global\savewidth\arrayrulewidth
  \global\arrayrulewidth 0.8pt}\hline\noalign{\global\arrayrulewidth\savewidth}}
\usepackage[pagebackref,breaklinks,colorlinks,allcolors=cvprblue]{hyperref}
\hypersetup{
    colorlinks=true,
    linkcolor=red, %IndianRed2,      % \ref 的颜色
    citecolor=LightBlue4, % blue,
}

\definecolor{cvprblue}{rgb}{0.21,0.49,0.74}
\definecolor{LightBlue4}{RGB}{104, 131, 139}
\definecolor{LightBlue1}{RGB}{191, 239, 255}
\definecolor{Bisque1}{RGB}{255, 228, 196}

\title{The Devil is in Temporal Token: High Quality Video Reasoning Segmentation} 

\author{Sitong Gong$^1$,
~Yunzhi Zhuge$^{1*}$,
~Lu Zhang$^1$,
~Zongxin Yang$^2$,\\
~Pingping Zhang$^1$,
~Huchuan Lu$^1$\\
\and
$^1$Dalian University of Technology, ~$^2$Havard University\\
}

\begin{document}
\maketitle
\begin{abstract}
Existing methods for Video Reasoning Segmentation rely heavily on a single special token to represent the object in the keyframe or the entire video, inadequately capturing spatial complexity and inter-frame motion. To overcome these challenges, we propose \textbf{VRS-HQ}, an end-to-end video reasoning segmentation approach that leverages Multimodal Large Language Models (MLLMs) to inject rich spatiotemporal features into hierarchical tokens. Our key innovations include a Temporal Dynamic Aggregation (TDA) and a Token-driven Keyframe Selection (TKS). Specifically, we design frame-level \SEG and temporal-level \TAK tokens that utilize MLLM’s autoregressive learning to effectively capture both local and global information. Subsequently, we apply a similarity-based weighted fusion and frame selection strategy, then utilize SAM2 to perform keyframe segmentation and propagation. To enhance keyframe localization accuracy, the TKS filters keyframes based on SAM2’s occlusion scores during inference. 
\textbf{VRS-HQ} achieves state-of-the-art performance on ReVOS, surpassing VISA  by  \textbf{5.9}\%/\textbf{12.5}\%/\textbf{9.1}\% in $\mathcal{J}$\&$\mathcal{F}$ scores across the three subsets. These results highlight the strong temporal reasoning and segmentation capabilities of our method. Code and model weights will be released at \href{https://github.com/SitongGong/VRS-HQ}{VRS-HQ}.
% Additionally, on the RVOS dataset, VRS-HQ surpasses the previous best method by an average of \textbf{x.x}% in $\mathcal{J}$&$\mathcal{F}$
% with \textbf{63.3}\%/\textbf{56.8}\%/\textbf{60.0}\% $\mathcal{J}$\&$\mathcal{F}$ on the ReVOS benchmark and \textbf{71.0}\%/\textbf{74.4}\%/\textbf{50.9}\% $\mathcal{J}$\&$\mathcal{F}$ on three standard RVOS datasets, demonstrating robust temporal reasoning capabilities of our approach. 
% Through the effective combination of these designs, our method achieves state-of-the-art performance on multiple video segmentation benchmarks and demonstrates competitive results on image segmentation datasets.
\end{abstract}    
\section{Introduction}
% \label{sec:intro}
% 陈述视频推理分割任务+论述一下之前方法的问题+我们方案的核心。
% 我们的工作主线为通过输入的文本驱动多模态大语言模型进行时序动态融合并提出了一种多条件自适应的选择算法来选择出对应于全局seg token的最合理的关键帧。
% 第一段首先引出视频推理分割任务，随着多模态大模型的发展，目前的推理分割方法局限于图像，视频推理分割方法最近被提出
% reasoning segmentation的意义，相比于传统的分割任务，
% In real-world applications, identifying the objects of interest from intricate natural language instructions has become essential for deep learning methods.
Reasoning segmentation~\cite{lai2024lisa,ren2024pixellm,rasheed2024glamm,he2024multi,xia2024gsva}, which aims to generate segmentation results from complex query texts, has advanced with multimodal foundation models~\cite{ravi2024sam, liu2024visual,li2023blip}. However, existing methods~\cite{lai2024lisa,ren2024pixellm,rasheed2024glamm,he2024multi} focus primarily on image-level segmentation,  leaving the more challenging domain of video-level segmentation, which requires temporal reasoning of object relations and attributes, relatively unexplored. To address this gap, Video Reasoning Segmentation (VRS)~\cite{yan2024visa,zheng2024villa,bai2024one} has recently emerged as a promising approach.
Unlike Referring Video Object Segmentation (RVOS) methods~\cite{wu2022language,wu2023onlinerefer,yan2024referred}, which rely on explicit descriptive phrases like ``a person skateboarding", VRS leverages the extensive world knowledge and temporal reasoning capabilities of Multimodal Large Language Models (MLLMs) to transform implicit intent-based expressions into precise object masklets.
% This advancement is critical for driving contemporary technologies, including autonomous driving and Embodied AI systems.

\begin{figure*}
    \centering
    \vspace{-3mm}
    \includegraphics[width=1\linewidth]{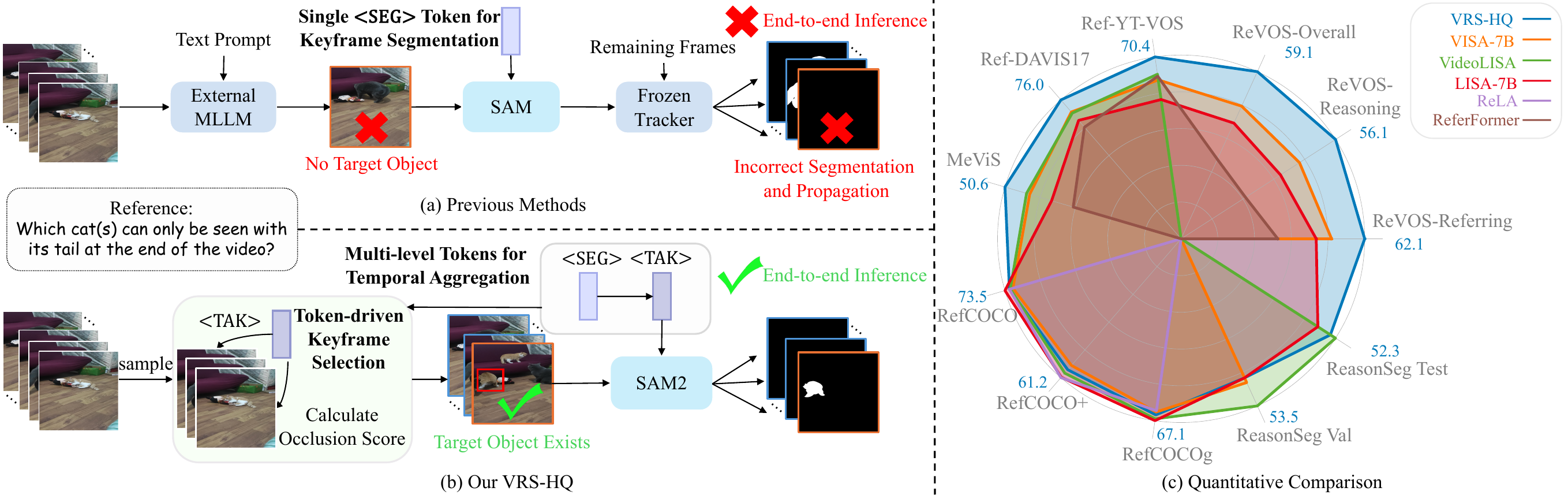}
    \vspace{-7mm}
    \caption{\textbf{Comparison with previous VRS approaches}.
    % (a) Previous methods rely heavily on an external MLLM to detect keyframes during inference, then perform keyframe-based single-frame segmentation using a single \SEG token, followed by mask propagation with a frozen tracker. This approach cannot perform end-to-end inference and suffers substantial segmentation loss if the MLLM misidentifies the keyframe.
    (a) Previous methods utilize a single \SEG token for keyframe-based segmentation, depending heavily on external models for keyframe detection and mask propagation. This reliance can hinder accurate keyframe localization and prevent end-to-end inference. 
    (b) In contrast, VRS-HQ introduces frame-level \SEG and a temporal \TAK token for dynamic aggregation. The aggregated \TAK token is then used for both keyframe selection and mask generation within SAM2. This enables single-stage inference with precise keyframe selection and high-quality segmentation.
    (c) VRS-HQ achieves state-of-the-art performance on various image and video datasets across reasoning and referring segmentation. 
    % Experiments validate that VRS-HQ achieves state-of-the-art performance across multiple image or video-based reasoning and referring segmentation datasets.
    }
    \label{fig:1}
    \vspace{-3mm}
\end{figure*}

\begin{figure*}
    \centering
    \vspace{-3mm}
    \includegraphics[width=1\linewidth]{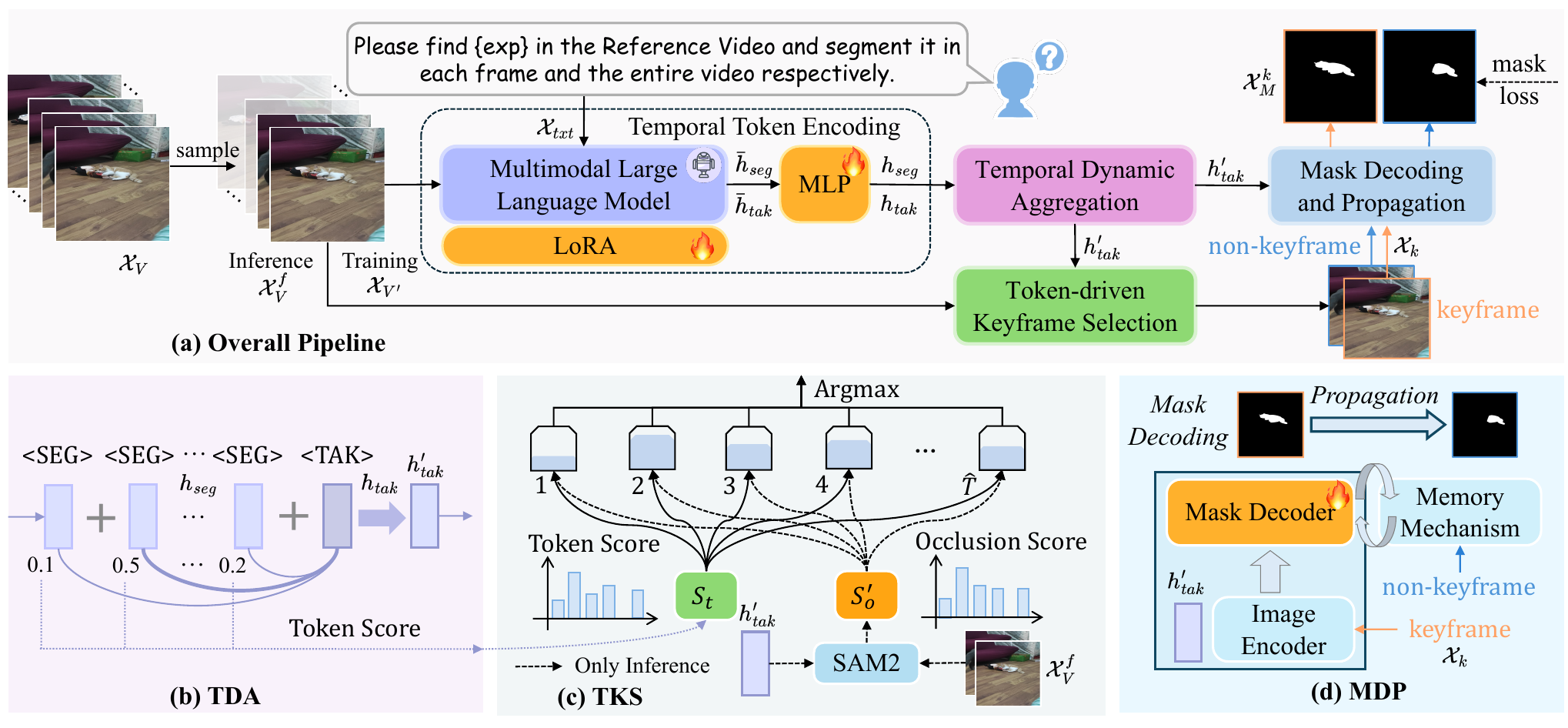}
    \vspace{-5mm}
    \caption{
    (a) \textbf{VRS-HQ architecture}. VRS-HQ incorporates a Multimodal Large Language Model for \textbf{Temporal Token Encoding} (\SEG and \TAK tokens, \textsection\ref{sec3.1}), a Temporal Dynamic Aggregation, a Token-driven Keyframe Selection and Mask Decoding and Propogation. 
    (b) \textbf{Temporal Dynamic Aggregation} (TDA) merges frame-level \SEG tokens into a temporal \TAK token using a weighted fusion based on cosine similarity. (\textsection\ref{sec3.2}). 
    (c) \textbf{Token-driven Keyframe Selection} (TKS). During training, the frame with the \SEG token closest to the \TAK token is selected as the keyframe. During inference, keyframe selection is refined using SAM2's occlusion scores and token similarity scores (\textsection\ref{sec3.3}).
    (d) \textbf{Mask Decoding and Propagation} (MDP). The \TAK token provides a sparse embedding for SAM2, generating a keyframe mask and propagating it to other frames via a memory mechanism (\textsection\ref{sec3.4}). 
    }
    % \hspace{0.15em}(\textsection\ref{sec3.0})
    \label{fig:2}
    \vspace{-3mm}
\end{figure*}

% 第二段说一下之前的方法主要聚焦于image-level，且图像，概括一下之前的方法及其局限性
% 重新改一下第二段，这次需要结合图1进行解释，同时要和abstract中的内容相对应，主要聚焦于之前的方法仅通过一个token进行单帧分割，不能提取视频级信息，而我们的核心是token聚合策略。另外主要提一下之前的方法选关键帧和传播过程均有局限性，而我们通过temporal token来选择关键帧同时充分利用SAM2强大的分割能力。
Despite recent advancements in Video Reasoning Segmentation (VRS), such as VISA~\cite{yan2024visa} and VideoLISA~\cite{bai2024one}, significant challenges still exist.
% While Video Reasoning Segmentation (VRS) has seen recent advancements with the introduction of various approaches such as VISA, ViLLa~\cite{zheng2024villa}, and VideoLISA~\cite{bai2024one}, these methods still face challenges in this field. 
\textbf{(i) Limited Temporal Context:}
Existing methods~\cite{yan2024visa,bai2024one} typically rely on a single segmentation token from an MLLM for keyframe-based segmentation (\cf~\textcolor{red}{Fig.}~\ref{fig:1} (a)), resulting in limited temporal context and hindering the effective capture of inter-frame variations and spatiotemporal features.
% As illustrated in \textcolor{red}{Fig.}~\ref{fig:1} (a), existing approaches~\cite{yan2024visa,bai2024one} typically employ a single \SEG token generated by Multimodal Large Langauge Model (MLLM) for keyframe-based single-frame or video-level segmentation, falling short in capturing inter-frame variations and effectively extracting spatiotemporal features due to the restricted representational capacity. 
\textbf{(ii) Suboptimal Keyframe Detection:}
The LLaMA-VID~\cite{li2025llama} model, used by VISA for keyframe detection, can produce inaccurate keyframes, particularly in videos requiring complex temporal reasoning. 
% leading to imprecise keyframe selection and hindering accurate mask propagation.
\textbf{(iii) Decoupled Segmentation and Propagation:} VISA's reliance on separate, pre-trained models for keyframe segmentation (SAM~\cite{kirillov2023segment}) and mask propagation (XMem~\cite{cheng2022xmem}) prevents end-to-end training and inference.
To address the above obstacles, we introduce several strategies to strengthen the VRS model’s proficiency in perceiving spatial information and interpreting temporal dynamics.
\textbf{Firstly}, current limitations arise from insufficient single-token representation capacity, which restricts the model's abilities to capture intra-frame spatial features and maintain inter-frame temporal coherence. 
We hypothesize that encoding frame-level and temporal-level information into hierarchical tokens separately via MLLM and integrating them could effectively unify spatial details with temporal dynamics, enhancing perceptual capability. 
% introducing and integrating hierarchical tokens could effectively encode frame-level and temporal-level information, 
% \textbf{Firstly}, we propose a hierarchical token representation that encodes frame-level and temporal-level information separately. By integrating these hierarchical tokens, we effectively combine spatial details within frames with temporal dynamics between frames, overcoming limitations of existing single-token representations.
\textbf{Secondly}, the occlusion score introduced by SAM2~\cite{ravi2024sam} inspires us to incorporate the target confidence of each sampled frame as a criterion for keyframe determination. 
We thus employ the temporal token in conjunction with SAM2 to generate the occlusion scores, applying temporal information for precise keyframe selection. 
As shown in \textcolor{red}{Fig.}~\ref{fig:1} (b), our approach locates the cat more accurately compared to VISA, boosting inference efficiency and segmentation performance.
% \textbf{Secondly},  to fully leverage temporal information for precise keyframe selection, we employ the temporal token in conjunction with SAM2 to generate occlusion scores as a basis for determining the keyframe, which reflects the target’s confidence in each frame.
\textbf{Thirdly}, leveraging SAM2's integrated segmentation and propagation capabilities, we efficiently fine-tune it with our temporal token and its memory mechanism for improved mask quality and inference efficiency.
In this work, we present \textbf{VRS-HQ} (\textbf{H}igh-\textbf{Q}uality \textbf{V}ideo \textbf{R}easoning \textbf{S}egmentation) (\cf~\textcolor{red}{Fig.}~\ref{fig:2}), a novel approach that leverages aggregated temporal tokens for enhanced keyframe selection and efficient mask decoding.
% as depicted in ~\ref{fig:2}, the main concept of which lies in harnessing the potential of the temporal token by aggregating them for enhanced keyframe selection and efficient mask decoding.
% The main concept of VRS-HQ is to integrate spatial features and temporal relations into special tokens to facilitate temporal dynamic aggregation, performing keyframe selection and mask decoding through the fused temporal token.
To begin with, we propose the \textbf{Temporal Token Encoding}(\textsection\ref{sec3.1}), prompting the MLLM to encode frame- and video-level target features into multi-level special tokens
% \TAK and \SEG 
using sampled video frames and tailored conversation templates.
% utilizing reference video frames and tailored conversation templates to prompt the MLLM to encode target features at both frame- and video-level into multi-level special tokens \TAK and \SEG.
The primary innovation lies in the \textbf{Temporal Dynamic Aggregation} (\textsection\ref{sec3.2}), which employs a cosine similarity-based weighted fusion strategy to merge frame-level \SEG embeddings into the temporal-level \TAK embedding, consolidating spatial features into \TAK while maintaining the temporal consistency of the targets.
% consolidating intra- and inter-frame spatiotemporal information. 
Subsequently, we select the keyframe according to the token similarity between \SEG and \TAK during training.  
% inputting it with the fused \TAK token into SAM2 for mask decoding.
Moreover, we propose the \textbf{Token-driven Keyframe Selection} (\textsection\ref{sec3.3}) during inference, sequentially treating each sampled frame as a potential keyframe and interacting it with fused \TAK embedding through SAM2 to calculate an occlusion score. 
The scores combined with previous token similarity are used as the criterion for keyframe selection.
% the reference frame with the \SEG token closest to the \TAK token as the keyframe during training, inputting it with the fused \TAK token into SAM2 for mask decoding. 
% Finally, the keyframe with the fused \TAK embedding is fed into SAM2 for mask decoding, while the remaining frames are regarded as non-keyframes and utilize SAM2’s memory mechanism for mask propagation (\textsection\ref{sec3.4}).
Finally, the keyframe with the fused \TAK embedding is fed into SAM2 for mask generation,  while the remaining frames are treated as non-keyframes and utilize SAM2's memory mechanism for \textbf{Mask Decoding and Propagation} (\textsection\ref{sec3.4}). 
% Finally, the adjacent frames are sampled as non-keyframes and fed into SAM2, leveraging its memory mechanism for mask propagation (\textsection\ref{sec3.4}). 
% enabling more effective fine-tuning (\textsection\ref{sec4.3}).
% Finally, we propose the \textbf{Token-driven Keyframe Selection} (\textsection\ref{sec3.3}) module during inference, sequentially treating each sampled reference frame as a potential keyframe and interacting it with \TAK embedding through SAM2 to calculate an occlusion score. The scores combined with token similarity are used as the criterion for keyframe identification, enhancing the accuracy of keyframe selection.
% Finally, we propose the \textbf{Token-driven Keyframe Selection} (\textsection\ref{sec4.4}) module to accurately identify the keyframe based on temporal information during inference. 
% To be specific, we use CLIP for global sampling and sequentially treat each reference frame as a potential keyframe, interacting it with \TAK embedding through SAM2 to calculate an occlusion score representing the target confidence for the current frame.
% The final keyframe is determined by combining token similarity with occlusion scores.

Through extensive experiments, \textbf{VRS-HQ} achieves state-of-the-art performance across a diverse range of video segmentation benchmarks (\cf~\textcolor{red}{Fig.}~\ref{fig:1} (c)). 
Specifically, on the ReVOS~\cite{yan2024visa} dataset, VRS-HQ outperforms VISA-13B by \textbf{9.1}\% in terms of the $\mathcal{J}$\&$\mathcal{F}$ metric, highlighting the critical role of our proposed module in enhancing the reasoning ability for scenes with complex spatiotemporal dynamics.
% the critical role of our proposed module in enhancing the model's reasoning ability towards scenes with complex spatial and temporal relationships.
Moreover, VRS-HQ surpasses previous methods by \textbf{7.3}\%/\textbf{5.6}\%/\textbf{6.5}\% $\mathcal{J}$\&$\mathcal{F}$ on three standard referring video segmentation datasets respectively, underscoring its strong inter-frame perception and robust video tracking capabilities.
Key contributions can be summarized as follows:
\begin{itemize}
    \item We present \textbf{Temporal Dynamic Aggregation} to blend spatial features from frame-level tokens into a temporal token, endowing the model with the ability to discern inter-frame variations and comprehend the global semantic context of the targets.
   %  We introduce VRS-HQ, which combines the video understanding capabilities of MLLM with the segmentation and propagation abilities of SAM2 to perform high-quality video reasoning segmentation.    
    \item We focus on harnessing the power of temporal tokens in video perception, using the integrated temporal token via SAM2 for keyframe segmentation and propagation. 
    Additionally, we present the \textbf{Token-driven Keyframe Selection}, combining each sampled frame with the temporal token to generate occlusion scores via SAM2, providing a reliable basis for keyframe detection.
    % serving as a robust criterion for accurate keyframe identification
    % significantly enhancing the model’s keyframe detection accuracy and global video context awareness.
    \item By combining the above designs, we introduce \textbf{VRS-HQ}, demonstrating state-of-the-art performance on the VRS benchmark and existing RVOS datasets.
\end{itemize}

\section{Related Works}
% \label{sec:formatting}

%-------------------------------------------------------------------------
\subsection{Referring Video Object Segmentation}
Referring Video Object Segmentation~\cite{wu2022language,botach2022end,yan2024referred} (RVOS), which focuses on segmenting and tracking prominent objects in video frames using explicit textual descriptions, has significantly advanced through the integration of visual and linguistic cues. A large proportion of these studies have leveraged attention mechanisms to merge multimodal information. For instance, Wang~\et~\cite{wang2019asymmetric} uses asymmetric cross-guided attention to enhance sentence representations and aggregate visual context.  Seo~\et~\cite{seo2020urvos} employ cross-modal and memory attention to jointly address RVOS and semi-supervised video object segmentation.  Hui~\et~\cite{hui2021collaborative} dynamically recombine linguistic features and interact with visual features using cross-modal attention to highlight spatiotemporal regions of interest.  Inspired by DETR~\cite{carion2020end},  methods like MTTR~\cite{botach2022end} and ReferFormer~\cite{wu2022language} leverage language queries for precise target localization. While RVOS excels with explicit object descriptions, it often struggles with the complex reasoning required to interpret more implicit and intricate language instructions, motivating our exploration of Video Reasoning Segmentation. 

% Wang~\et~\cite{wang2019asymmetric} proposes an asymmetric cross-guided attention network that enhances sentence representation and aggregates query-focused visual context through vision and language-guided modules, respectively. Seo~\et~\cite{seo2020urvos} employs cross-modal and memory attention modules to frame the referring video object segmentation task as a joint endeavor with semi-supervised video object segmentation, thus enabling the simultaneous resolution of both tasks. Hui~\et~\cite{hui2021collaborative} proposes a framework that dynamically recombines linguistic features and interacts with visual features using cross-modal attention, effectively highlighting the correct actor's spatial and temporal regions. Drawing inspiration from the DETR~\cite{carion2020end}, MTTR~\cite{botach2022end}, and ReferFormer~\cite{wu2022language} have been developed, utilizing language as queries to enable a more precise localization to the referring target.

%-------------------------------------------------------------------------
\subsection{Reasoning Segmentation}
Reasoning segmentation~\cite{lai2024lisa,ren2024pixellm,rasheed2024glamm,xia2024gsva} advances referring segmentation by generating masks from complex images and versatile text prompts.  
% This requires models to possess cross-modal comprehension abilities to interpret complex visual scenes and natural-language cues effectively. 
LISA~\cite{lai2024lisa} pioneers the field of reasoning segmentation. It introduces a new token to expand the vocabulary and proposes the embedding-as-mask paradigm, enhancing segmentation capabilities to address scenarios requiring complex reasoning and world knowledge. LISA++~\cite{yang2023improved} addresses LISA's limitations in distinguishing individual instances by enhancing instruction-tuning data with key segmentation datasets, supporting both semantic and instance segmentation tasks. PixelLM~\cite{ren2024pixellm} innovatively combines a novel pixel decoder and a segmentation codebook with learnable tokens to efficiently generate high-quality masks without external models. VISA~\cite{yan2024visa} introduces reasoning segmentation to the video domain, proposing reasoning video object segmentation. It uses a pre-trained Multimodal Large Language Model (MLLM) to select keyframes, segments them based on reasoning, and propagates the masks to other frames using a pre-trained object tracker. However, it suffers from limited representation capability of a single special token, inaccurate keyframe selection, and the inability to perform end-to-end inference, compromising its segmentation and tracking performance. 

\vspace{-1mm}
\section{Methodology}
\label{method}
\vspace{-1mm}
\paragraph{Task Defination.}
The task of Video Reasoning Segmentation can be briefly outlined as follows.
Given a video clip consisting of $T$ frames $\mathcal{X}_{V}\in{\mathbb{R}^{T\times{3}\times{H}\times{W}}}$, where $H$ and $W$ denote the height and width of each frame respectively, along with a high-level textual instruction $\mathcal{X}_{txt}$, 
%incorporating human intent and world knowledge,
VRS aims to design a model $\mathcal{M}$ to interpret and transform $\mathcal{X}_{txt}$ into the binary segmentation mask sequence $\mathcal{X}_{M}\in{\mathbb{R}^{T\times{H}\times{W}}}$ for each frame. 
% \vspace{-2mm}
% \begin{equation}
% \mathcal{X}_{M}=\mathcal{M}(\mathcal{X}_{V}, \mathcal{X}_{txt})
% \end{equation}
% \vspace{-7mm}
In contrast to RVOS tasks providing explicit descriptions like ``the person skateboarding," VRS typically employs expressions incorporating world knowledge like ``tool(s) for holding garbage" or temporal logic like ``the ship(s) moving at the highest speed.", raising higher demands on the model's capabilities of temporal relationship comprehension and complex scenario reasoning.
\vspace{-4mm}
\paragraph{Overall Architecture.}
\label{sec3.0}
\textcolor{red}{Fig.}~\ref{fig:2} illustrates the VRS-HQ architecture, which comprises Chat-UniVi~\cite{jin2024chat} as the Multimodal Large Language Model (MLLM) for temporal token encoding (\textsection\ref{sec3.1}), a Temporal Dynamic Aggregation (\textsection\ref{sec3.2}), a Token-driven Keyframe Selection (\textsection\ref{sec3.3}), and Mask Decoding and Propagation (\textsection\ref{sec3.4}).  The MLLM encodes the textual prompt into multi-level tokens representing spatial and semantic target information. These tokens drive both temporal dynamic aggregation and keyframe selection via cosine similarity scores.  SAM2 then segments the keyframes and tracks the target object throughout the video.  During inference, the Token-driven Keyframe Selection computes object occlusion scores for each sampled frame by merging it with the fused temporal token. Filtering keyframes based on these scores improves the accurate localization of the target object and subsequently segmentation process.

\subsection{Temporal Token Encoding}
\label{sec3.1}
% 首先进行动机阐述，之前的方法均适用单一token进行分割而具有局限性，我们设计了两种token分别代表不同层级的信息。
Current VRS approaches~\cite{yan2024visa,bai2024one} typically use specified language prompts to guide MLLMs in embedding target information from keyframes or entire videos into a single specialized token, struggling to capture rich spatiotemporal dynamics essential for fine-grained video understanding.  
To equip the MLLM with both frame-level and video-level contextual awareness for video segmentation, we propose encoding intra-frame spatial information and inter-frame temporal relations into hierarchical tokens.  
\vspace{-4mm}
\paragraph{Hierarchical Token Generation.}
Instead of previous single-token encoding strategy, the vocabulary of the MLLM is initially augmented with two new special tokens: \SEG and \TAK\!. 
Then, we design a structured conversational template as ``\texttt{\textbf{USER}: Please find \{expression\} in the Reference Video and segment it in each frame and the entire video respectively.}"  Here, ``\{expression\}" denotes the target object description. 
The tokenized prompts $\mathcal{X}_{txt}$ and sampled video frames  $\mathcal{X}_{V'}$ are input into the MLLM, which generates a response ${y}_{txt}$ containing multiple frame-level \SEG tokens and a temporal-level \TAK token through autoregressive encoding.
% Current VRS approaches~\cite{yan2024visa,bai2024one} typically employ crafted language prompts to guide MLLMs in embedding target information from keyframes or entire videos into a single specialized token.  Such representations often prove inadequate for capturing the rich spatiotemporal dynamics essential for accurate video understanding.
% To address this, we introduce two hierarchical tokens, \SEG and \TAK\!, encoding intra-frame spatial information and inter-frame temporal relations, respectively.  Our approach integrates these tokens to provide a richer representation of temporal context.
% 具体做法
% To endow the MLLM with both local and global perception and segmentation proficiency for video, 
\vspace{-4mm}
\paragraph{Token Extraction and Mapping.}
Subsequently, the \SEG token embeddings $\bar{h}_{seg} \in \mathbb{R}^{T' \times d'}$ and the \TAK token embedding $\bar{h}_{tak} \in \mathbb{R}^{1 \times d'}$ are extracted from the MLLM's final layer, where $d'$ denotes the MLLM's embedding dimension and $T'$ denotes the length of sampled frames. These embeddings are then projected into the same feature space as SAM2 using a multi-layer perceptron:
\vspace{-1mm}
\begin{equation}
h_{seg}, h_{tak}=MLP(\bar{h}_{seg}), MLP(\bar{h}_{tak})
\end{equation}
% \vspace{-1=2mm}
Here, $h_{seg} \in \mathbb{R}^{T' \times d}$ and $h_{tak} \in \mathbb{R}^{1 \times d}$ represent the sparse embeddings for segmentation mask activation, and $d$ denotes the feature dimension of SAM2. 
Finally, the frame-level embeddings $h_{seg}$ are aggregated with the video-level embedding $h_{tak}$ to incorporate temporal dynamics.

% where $MLP$ represents the MLP layer, $h_{seg}\in{\mathbb{R}^{T'\times{d}}}$ and $h_{tak}\in{\mathbb{R}^{1\times{d}}}$ represent the final obtained sparse embeddings for segmentation masks activation, and $d$ denotes the feature dimension of SAM2. 
% After extracting the token embeddings, we merge the frame-level $h_{seg}$ with the video-level $h_{tak}$, effectively aggregating temporal dynamics across frames.

\subsection{Temporal Dynamic Aggregation}
\label{sec3.2}
% During the autoregressive processing of the MLLM, the \SEG token encapsulates the local features of the object within a single frame, while the \TAK token contains the object’s semantic and temporal information across the entire video.
% 动机：之前的ref-vos方法利用视频-帧级融合取得了好的效果。
% Inspired by recent advancements~\cite{luo2024soc,yan2024referred} in Ref-VOS, incorporating temporal embedding with local embeddings provides both spatial prior and temporal signals of the targets for the mask decoding. 
% Building on recent advances~\cite{luo2024soc,yan2024referred} in Ref-VOS, incorporating temporal with local embeddings provides both spatial prior and temporal signals of the targets for the mask decoding. 
Building on the strong temporal encoding of MLLM, \SEG and \TAK embeddings encapsulate the spatial priors and temporal semantic signals of the targets respectively, providing rich contextual information for the segmentation model. 
Accordingly, we propose the Temporal Dynamic Aggregation to facilitate the fusion of positional and semantic information of the targets.
% integrate both to enhance inter-frame consistency while achieving more precise discernment of the object’s positional shifts across frames. 
% Inspired by prior works~\cite{lai2024lisa,bai2024one}, through the autoregressive processing of the MLLM, local embedding $h_{seg}$ implicitly contains spatial position information of the target across frames, while global embedding $h_{tak}$ integrates the MLLM’s interpretation of its semantic content.
% Thus, we propose combining them to enhance inter-frame consistency while more accurately capturing the object's positional transforms across frames. 
% Accordingly, we advocate for their integration to strengthen inter-frame coherence and achieve more precise discernment of the object’s positional shifts across frames
\vspace{-4mm}
\paragraph{Keyframes as Token Similarity.} The cosine similarity between each frame-level \SEG token and the temporal-level \TAK token reflects the semantic alignment between individual frames and the overall video context.  We hypothesize that high-similarity frames are more representative of the video's content and thus suitable as keyframes. This motivates our use of token similarity for keyframe selection during training, which also facilitates weighted fusion of \SEG and \TAK tokens.
% Based on the analysis of the hierarchical tokens, we suppose that the cosine similarity between each \SEG token and the temporal \TAK token represents the semantic distance of the targets across individual frames and the global video.
% Therefore, we consider using this similarity as the criterion for selecting keyframes during training, leveraging it to perform the weighted fusion of \SEG and \TAK tokens simultaneously.

\vspace{-4mm}
\paragraph{Similarity-based Weighted Fusion. }
% To be specific, we initially calculate the cosine similarity between $h_{tak}$ and $h_{seg}$ of each frame, measuring the representational relevance of the target across each sampled frame and the global video. 
To enhance the inter-frame consistency while achieving more precise discernment of the object’s positional shifts across frames, 
we fuse the frame-level embeddings $h_{seg}$ with the video-level embedding $h_{tak}$ using normalized cosine similarity scores as attention weights (\cf~\textcolor{red}{Fig.}~\ref{fig:2} (b)). During backpropagation, frame-level features $h_{seg}$ are updated alongside the temporal-level embedding $h_{tak}$ to enable more comprehensive learning of local and global video context:
\begin{equation}
\label{eq:temporal aggregation}
h'_{tak} = h_{tak} + \alpha \sum_{i=1}^{T'} \lambda_{i} h_{seg}[i],
\end{equation}
where $\lambda_{i}$ represents the normalized cosine similarity between $h_{seg}[i]$ and $h_{tak}$, and $\alpha$ is the fusion coefficient. The resulting fused embedding $h'_{tak}$ is then used for keyframe selection during inference and mask generation.

% Then we utilize the normalized similarity scores as the attention weights to fuse $h_{seg}$ into $h_{tak}$ (\cf~\textcolor{red}{Fig.}~\ref{fig:2} (b)). During gradient backpropagation, $h_{seg}$ can be updated and optimized alongside $h_{tak}$, enabling more comprehensive learning of both local and global video context,
% the process of which can be formulated as:
% \begin{equation}
% \label{eq:temporal aggregation}
% h'_{tak} = h_{tak} + \alpha\sum^{T}_{i=1}{\lambda_{i}h_{seg}[i]}
% \end{equation}
% where $\lambda_{i}$ represents the normalized cosine similarity scores between $h_{seg}$ and $h_{tak}$ and $\alpha$ represents the fusion coefficient.  
% The fused embedding $h'_{tak}$ is used for keyframe mask generation and propagation through SAM2.

% \begin{figure}
%     \centering
%     \vspace{-3mm}
%     \includegraphics[width=0.9\linewidth]{author-kit-CVPR2025-v3.1-latex-/Figures/TKS.pdf}
%     \vspace{-3mm}
%     \caption{Token-driven Keyframe Selection module (\textsection\ref{sec4.4}).
%     % The TKS module sequentially treats each reference frame as a keyframe, using SAM2 and the fused \TAK token to generate an occlusion score for each frame. 
%     The TKS module sequentially generates an occlusion score for each reference frame using the merged \TAK token via SAM2. 
%     The occlusion score is then combined with the cosine similarity calculated in the TDA module to determine the keyframe. 
%     }
%     \label{TKS}
%     \vspace{-3mm}
% \end{figure}

\subsection{Token-driven Keyframe Selection}
\label{sec3.3}
% 动机：之前的方法依赖于复杂的提示设计来查找关键帧。
VISA~\cite{yan2024visa} relies on an external model (LLaMA-VID~\cite{li2025llama}) for keyframe selection during inference, hindering end-to-end processing and potentially degrading performance due to inaccuracies in the external model's output. To address this limitation, we introduce the Token-driven Keyframe Selection that leverages the temporal information encoded within the integrated \TAK embedding. This approach eliminates the need for complex prompt engineering and significantly improves the reliability of keyframe selection.

Instead of uniform video sampling during training, we adopt the CLIP model~\cite{radford2021learning} to find the frame most aligned with the expression $\mathcal{X}_{exp}$ for inference. 
The anchor frame is used for global sampling, 
%with frame with the highest similarity serves as the anchor point for global sampling, 
resulting in sampled frames $\mathcal{X}^{f}_{V}\in{\mathbb{R}^{\hat{T}\times{3}\times{H}\times{W}}}$, where $\hat{T}$ represents the number of the sampled frames. 
Following the temporal dynamic aggregation (\textcolor{red}{Eq.}~\ref{eq:temporal aggregation}), 
each sampled frame is treated as a potential keyframe, along with the fused \TAK embedding $h'_{tak}$, is sent to SAM2 to generate the object occlusion scores $S_{o}\in{\mathbb{R}^{\hat{T}\times{1}}
}$ (\cf~\textcolor{red}{Fig.}~\ref{fig:2} (c)):
\begin{equation}
S_{o} = \mathcal{MD}(\mathcal{E}(\mathcal{X}^{f}_{V}), h'_{tak})
\end{equation}
The occlusion scores effectively indicate the confidence of the target’s presence in the current frame, assisting in identifying which frame best corresponds to the information encoded in the \TAK token. 
After determining the keyframe index by the combination of softmax-normalized scores $S'_{o}$ with token similarity scores $S_{t}$, we decode the keyframe mask and then utilize the cross-frame propagation technique of SAM2 to obtain video-level segmentation masks.

\subsection{Mask Decoding and Propagation}
\label{sec3.4}
% Previous ReasonVOS methods~\cite{zheng2024villa,yan2024visa,bai2024one} have all utilized perception models pre-trained on image datasets, such as SAM, to perform fine-grained on single frames or entire videos.
% 动机：之前的方法采用外部模型或图像级模型进行视频级追踪。
Following temporal token fusion and keyframe selection, the mask embeddings enriched with positional and semantic information can be yielded by the segmentation model. 
In contrast to prior methods~\cite{zheng2024villa,yan2024visa,bai2024one} depending on image-level segmentation models or external object trackers for target trajectories prediction, we utilize SAM2 to perform segmentation and propagation concurrently. 
\vspace{-3mm}
\paragraph{SAM2 for Masklets Decoding.}
Given the keyframe $\mathcal{X}_{k}$, we first extract its features using the image encoder $\mathcal{E}$, providing the conditional input for SAM2.  The integrated temporal embedding $h'_{tak}$ then interacts with these keyframe features within the mask decoder $\mathcal{MD}$ (\cf~\textcolor{red}{Fig.}~\ref{fig:2} (d)) to generate the keyframe mask:
\begin{equation}
\label{equ:mask decoding}
\mathcal{X}^{k}_{M}=\mathcal{MD}(\mathcal{E}(\mathcal{X}_{k}),h'_{tak}),
\end{equation}
where $\mathcal{X}^{k}_{M}$ represents the predicted mask for the keyframe.  Next, we propagate this mask to two adjacent frames, $\mathcal{X}_{k-1}$ and $\mathcal{X}_{k+1}$, treated as non-conditional frames, using the memory storage and interaction mechanism. This yields the mask sequence $\mathcal{X}^{s}_{M}$. During inference, all remaining frames are processed as non-conditional frames, facilitating mask propagation throughout the entire video.

% In parallel with SAM2’s video training strategy, the two additional sampled frames are treated as non-conditional frames. Using the memory storage and interaction mechanism, the keyframe mask is propagated to these frames, culminating in the mask sequence $\mathcal{X}^{s}_{M}$. 
% using the mask memory mechanism to propagate the keyframe mask to these frames, culminating in the mask sequence $ \mathcal{X}^{s}_{M}$ of these frames. 
% Following SAM2’s video training strategy, the two additional sampled frames are treated as non-conditional frames. Using the mask memory mechanism, the keyframe mask is propagated to these frames, resulting in the mask sequence \textbackslash{}mathcal\{X\}^\{s\}_\{M\}.
% \vspace{-4mm}

\subsection{Training Objectives}
Our method is trained end-to-end using a combined text generation loss and mask loss to optimize the \TAK and \SEG embeddings. The mask loss combines binary cross-entropy (BCE) and DICE loss:
\begin{equation}
L_{total}=\lambda_{txt}L_{txt}({y}_{txt},\hat{y}_{txt})+\lambda_{mask}L_{mask},
\end{equation}
\begin{equation}
L_{mask}=\lambda_{bce}L_{bce}(\mathcal{X}^{s}_{M},\mathcal{\hat{X}}^{s}_{M})+\lambda_{dice}L_{dice}(\mathcal{X}^{s}_{M},\mathcal{\hat{X}}^{s}_{M}),
\end{equation}
where $\hat{y}_{txt}$ and $\mathcal{\hat{X}}^{s}_{M}$ represent the ground truth text and mask, respectively, and ${y}_{txt}$ and ${X}^{s}_{M}$ are their corresponding predictions. The weighting coefficients $\lambda_{txt}$, $\lambda_{mask}$, $\lambda_{bce}$, and $\lambda_{dice}$ are set to 1, 1, 2 and 0.5, respectively.

\section{Experiments}

\begin{table*}[htbp]
\centering
\vspace{-2mm}
\caption{Performance comparison with previous methods on ReVOS dataset.}
 % (\textsection\ref{Comparison Results})
% The comparison results of other methods are provided by VISA~\cite{yan2024visa}. }
\label{ReasonVOS}
\vspace{-3mm}
\scalebox{0.94}{
\begin{tabular}{rl||l||ccc|ccc|ccc|c}
% \toprule[1.2pt]
\shline
\rowcolor{gray!35}~& & & \multicolumn{3}{c|}{referring} & \multicolumn{3}{c|}{reasoning} & \multicolumn{3}{c|}{overall} &  \\
% \multirow{2}{*}{$\mathcal{R}$} \\
% \cline{4-12} % \cline{4-12}
% \multicolumn{2}{c||}{\multirow{-2}{*}{Methods}} 
\rowcolor{gray!35}\multicolumn{2}{c||}{\multirow{-2}{*}{Methods}} & \multirow{-2}{*}{Backbone} & $\mathcal{J}$ & $\mathcal{F}$ & $\mathcal{J}\&\mathcal{F}$ & $\mathcal{J}$ & $\mathcal{F}$ & $\mathcal{J}\&\mathcal{F}$ & $\mathcal{J}$ & $\mathcal{F}$ & $\mathcal{J}\&\mathcal{F}$ &\multirow{-2}{*}{$\mathcal{R}$}\\
\shline
   \multicolumn{1}{r}{MTTR$_{\!}$~\cite{botach2022end}}&\!\!\!\pub{CVPR2022}\!\!   & Video-Swin-T & 29.8 & 30.2 & 30.0 & 20.4 & 21.5 & 21.0 & 25.1 & 25.9 & 25.5 & 5.6 \\
  \multicolumn{1}{r}{LMPM$_{\!}$~\cite{ding2023mevis}}&\!\!\!\pub{ICCV2023}\!\!    & Swin-T & 29.0 & 39.1 & 34.1 & 13.3 & 24.3 & 18.8 & 21.2 & 31.7 & 26.4 & 3.2 \\
  \multicolumn{1}{r}{ReferFormer$_{\!}$~\cite{wu2022language}}& \!\!\!\pub{CVPR2022}\!\!  & Video-Swin-B & 31.2 & 34.3 & 32.7 & 21.3 & 25.6 & 23.4 & 26.2 & 29.9 & 28.1 & 8.8 \\
  \multicolumn{1}{r}{LISA$_{\!}$~\cite{lai2024lisa}}& \!\!\!\pub{CVPR2024}\!\! & LLaVA-7B & 44.3 & 47.1 & 45.7 & 33.8 & 38.4 & 36.1 & 39.1 & 42.7  & 40.9 & 9.3 \\
  \multicolumn{1}{r}{LISA$_{\!}$~\cite{lai2024lisa}}& \!\!\!\pub{CVPR2024}\!\! & LLaVA-13B & 45.2 & 47.9 & 46.6 & 34.3 & 39.1 & 36.7 & 39.8 & 43.5 & 41.6 & 8.6 \\
  \multicolumn{1}{r}{TrackGPT$_{\!}$~\cite{zhu2023tracking}}&   \!\!\!\pub{arXiv2023}\!\! & LLaVA-7B & 46.7 & 49.7 & 48.2 & 36.8 & 41.2 & 39.0 & 41.8 & 45.5 & 43.6 & 11.6 \\
 \multicolumn{1}{r}{TrackGPT$_{\!}$~\cite{zhu2023tracking}}&   \!\!\!\pub{arXiv2023}\!\!  & LLaVA-13B &48.3 & 50.6 & 49.5 & 38.1 & 42.9 & 40.5 & 43.2 & 46.8 & 45.0 & 12.8 \\
  \multicolumn{1}{r}{{VISA}$_{\!}$~\cite{yan2024visa}}&  \!\!\!\pub{ECCV2024}\!\!     & LLaVA-7B & 49.4 & 52.6 & 51.0 & 40.5 & 45.8 & 43.2 & 44.9 & 49.2 & 47.1 & 15.3 \\
\multicolumn{1}{r}{{VISA}$_{\!}$~\cite{yan2024visa}}&  \!\!\!\pub{ECCV2024}\!\! & LLaVA-13B &55.7 &59.0 & 57.4 & 41.9 & 46.5 & 44.2 & 48.8 & 52.8 & 50.8 & 15.1 \\
  \multicolumn{1}{r}{{VISA}$_{\!}$~\cite{yan2024visa}}&  \!\!\!\pub{ECCV2024}\!\! & Chat-UniVi-7B &49.2 & 52.6 & 50.9 & 40.6 & 45.4 & 43.0 & 44.9 & 49.0 & 46.9 & 15.5\\
  \multicolumn{1}{r}{{VISA}$_{\!}$~\cite{yan2024visa}}&  \!\!\!\pub{ECCV2024}\!\! & Chat-UniVi-13B & 55.6 & 59.1 & 57.4 & 42.0 & 46.7 & 44.3 & 48.8 & 52.9 & 50.9 & 14.5\\
    \hline
   \multicolumn{1}{r}{VRS-HQ}&\!\!\!\pub{Ours}\!\!  & Chat-UniVi-7B &\second{59.8} & \second{64.5} & \second{62.1} & \second{53.5} & \second{58.7} & \second{56.1} & \second{56.6} & \second{61.6} & \second{59.1} & \best{19.7}\\
  \multicolumn{1}{r}{VRS-HQ}&\!\!\!\pub{Ours}\!\! & Chat-UniVi-13B & \best{61.1} & \best{65.5} &	\best{63.3} & \best{54.1} & \best{59.4} & \best{56.8} & \best{57.6} & \best{62.5} & \best{60.0} & \second{18.9} \\
% \bottomrule[1.2pt]
\shline
\end{tabular}}
% \vspace{-1mm}
\end{table*}

\subsection{Datasets and Metrics}
\paragraph{Datasets.}
Our model is trained and evaluated on extensive image and video segmentation datasets, LLaVA-Instruct-150k~\cite{liu2024visual}, as well as the Video Question-Answering datasets from Video-ChatGPT~\cite{maaz2023video}. 
Specifically, the image segmentation datasets comprise semantic segmentation: ADE20K~\cite{zhou2017scene}, COCO-Stuff~\cite{caesar2018coco}, PACO-LVIS~\cite{ramanathan2023paco}, and PASCALPart~\cite{chen2014detect}, referring segmentation: refCLEF, refCOCO, refCOCO+~\cite{kazemzadeh2014referitgame}, and refCOCOg~\cite{mao2016generation}, and reasoning segmentation: ReasonSeg~\cite{lai2024lisa}. 
While the video segmentation datasets encompass the RVOS datasets: Ref-Youtube-VOS~\cite{seo2020urvos}, Ref-DAVIS17~\cite{pont20172017} and MeViS~\cite{ding2023mevis}, and the VRS benchmark ReVOS~\cite{yan2024visa}.
\vspace{-4mm} 
\paragraph{Evaluation Metrics.}
We evaluate video segmentation using region similarity ($\mathcal{J}$), contour accuracy ($\mathcal{F}$), and their mean ($\mathcal{J}\&\mathcal{F}$).  Image segmentation accuracy is measured via Generalized Intersection over Union (gIoU)~\cite{rezatofighi2019generalized} and Complete Intersection over Union (cIoU)~\cite{zheng2020distance}. Model hallucination is assessed using the robustness score 
$\mathcal{R}$~\cite{yan2024visa}.

% Following previous methods~\cite{lai2024lisa,yan2024visa}, we use region similarity ($\mathcal{J}$), contour accuracy ($\mathcal{F}$), and their average value ($\mathcal{J}\&\mathcal{F}$) for the evaluation of video segmentation. While gIoU and cIoU are adopted to measure the accuracy of image segmentation.
% As for the evaluation of model hallucination, we employ the robustness score $\mathcal{R}$ following VISA~\cite{yan2024visa}. 

\begin{table*}[htbp]
\centering
\vspace{-1mm}
\caption{Performance comparison with previous methods on the validation sets of RVOS datasets.}
% The outcomes of LISA and TrackGPT are presented by VISA~\cite{yan2024visa}}
\label{RefVOS}
\vspace{-3mm}
\scalebox{0.99}{
\begin{tabular}{rl||l||ccc|ccc|ccc}
\shline
% \multirow{2}{*}{Method} 
%\multicolumn{2}{c||}{\multirow{-2}{*}{Methods}}
\rowcolor{gray!35} & &  & \multicolumn{3}{c|}{Ref-YouTube-VOS} & \multicolumn{3}{c|}{Ref-DAVIS17} & \multicolumn{3}{c}{MeViS} \\
% \cline{4-12}
\rowcolor{gray!35} \multicolumn{2}{c||}{\multirow{-2}{*}{Methods}}& \multirow{-2}{*}{Backbone} & $\mathcal{J}$ & $\mathcal{F}$ & $\mathcal{J}\&\mathcal{F}$ & $\mathcal{J}$ & $\mathcal{F}$ & $\mathcal{J}\&\mathcal{F}$ & $\mathcal{J}$ & $\mathcal{F}$ & $\mathcal{J}\&\mathcal{F}$\\
\shline
    \multicolumn{1}{r}{MTTR$_{\!}$~\cite{botach2022end}}&\!\!\!\pub{CVPR2022}\!\! & Video-Swin-T & 54.0 & 56.6 & 55.3 & - & - & - & 28.8 & 31.2 & 30.0 \\
    \multicolumn{1}{r}{LMPM$_{\!}$~\cite{ding2023mevis}}&\!\!\!\pub{ICCV2023}\!\!  & Swin-T & - & - & - & - & - & - & 34.2 & 40.2 & 37.2 \\
    \multicolumn{1}{r}{ReferFormer$_{\!}$~\cite{wu2022language}}& \!\!\!\pub{CVPR2022}\!\!  & Video-Swin-B & 61.3 & 64.6 & 62.9 & 58.1 & 64.1 & 61.1 & 29.8 & 32.2 & 31.0 \\
    \multicolumn{1}{r}{OnlineRefer$_{\!}$~\cite{wu2023onlinerefer}}& \!\!\!\pub{CVPR2023}\!\! & Swin-L &61.6 & 65.5 & 63.5 & 61.6 & 67.7 & 64.8& - & - & - \\
    % SgMg~\cite{miao2023spectrum} & Video-Swin-B~\cite{liu2022video} & 63.9 & 67.4 & 65.7 & 60.6 & 66.0 & 63.3 & - & - & - \\
    % SOC~\cite{luo2024soc} & Video-Swin-B~\cite{liu2022video} & 64.1 & 67.9 & 66.0 & 61.0 & 67.4 & 64.2 & - & - & - \\
    % AL-Ref-SAM2~\cite{huang2024unleashing} & - & 65.9 & 69.9 & 67.9 & 70.4 & \underline{78.0} & 74.2 & 39.5 & 46.2 & 42.8 \\
    \multicolumn{1}{r}{LISA$_{\!}$~\cite{lai2024lisa}}& \!\!\!\pub{CVPR2024}\!\! & LLaVA-7B & 53.4 & 54.3 & 53.9 & 62.2 & 67.3 & 64.8 & 35.1 & 39.4 & 37.2 \\
    \multicolumn{1}{r}{LISA$_{\!}$~\cite{lai2024lisa}}& \!\!\!\pub{CVPR2024}\!\!  & LLaVA-13B & 54.0 & 54.8 & 54.4 & 63.2 & 68.8 & 66.0 & 35.8 & 40.0 & 37.9 \\
    \multicolumn{1}{r}{TrackGPT$_{\!}$~\cite{zhu2023tracking}}&   \!\!\!\pub{arXiv2023}\!\! & LLaVA-7B & 55.3 & 57.4 & 56.4 & 59.4 & 67.0 & 63.2 & 37.6 & 42.6 & 40.1 \\
    \multicolumn{1}{r}{TrackGPT$_{\!}$~\cite{zhu2023tracking}}&   \!\!\!\pub{arXiv2023}\!\! & LLaVA-13B & 58.1 & 60.8 & 59.5 & 62.7 & 70.4 & 66.5 & 39.2 & 43.1 & 41.2 \\
    \multicolumn{1}{r}{{VISA}$_{\!}$~\cite{yan2024visa}}&  \!\!\!\pub{ECCV2024}\!\! & Chat-UniVi-7B & 59.8 & 63.2 & 61.5 & 66.3 & 72.5 & 69.4 & 40.7 & 46.3 & 43.5 \\
    \multicolumn{1}{r}{{VISA}$_{\!}$~\cite{yan2024visa}}&  \!\!\!\pub{ECCV2024}\!\! & Chat-UniVi-13B & 61.4 & 64.7 & 63.0 & 67.0 & 73.8 & 70.4 & 41.8 & 47.1 & 44.5 \\
    \multicolumn{1}{r}{VideoLISA$_{\!}$~\cite{bai2024one}}&  \!\!\!\pub{NeurIPS2024}\!\! & LLaVA-Phi-3-V & 61.7 & 65.7 & 63.7 & 64.9 & 72.7 & 68.8 & 41.3 & \underline{47.6} & 44.4 \\
    \hline
    \multicolumn{1}{r}{VRS-HQ}&  \!\!\!\pub{Ours}\!\!  & Chat-UniVi-7B & \second{68.3} & \second{72.5} &
    \second{70.4} & \best{72.6} & \best{79.4} & \best{76.0}& \second{47.6} & \best{53.7} & \second{50.6}\\
    \multicolumn{1}{r}{VRS-HQ}&  \!\!\!\pub{Ours}\!\!  & Chat-UniVi-13B & \best{69.0} & \best{73.1} & \best{71.0} & \second{71.0} & \second{77.9} & \second{74.4} & \best{48.0} & \best{53.7} & \best{50.9}\\
\shline
\end{tabular}
}
\vspace{-2mm}
\end{table*}
\begin{table*}[htbp]
\centering
\vspace{-2mm}
\caption{Performance comparison with previous methods on referring and reasoning image segmentation datasets.}
\label{RefCOCO}
\vspace{-3mm}
\scalebox{0.9}{
\begin{tabular}{l||l||ccc|ccc|cc|cc|cc}
  % 设置线宽为2pt
\shline
% \toprule[1.2pt]
\rowcolor{gray!35}  &  & \multicolumn{3}{c|}{refCOCO} & \multicolumn{3}{c|}{refCOCO+} & \multicolumn{2}{c|}{refCOCOg} & \multicolumn{2}{c|}{ReaSeg (val)} & \multicolumn{2}{c}{ReaSeg (test)} \\
% \cline{3-14}
\rowcolor{gray!35} \multirow{-2}{*}{Method}& \multirow{-2}{*}{Backbone}& val & testA & testB & val & testA & testB & val & test & gIoU & cIoU & gIoU & cIoU \\
% \midrule
\shline
    % MCN~\cite{luo2020multi}  & DarkNet53 & 62.4 & 64.2 & 59.7 & 50.6 & 55.0 & 44.7 & 49.2 & 49.4 & - & - & - & - \\
    % VLT~\cite{ding2021vision}  & DarkNet53 & 65.7 & 68.3 & 62.7 & 55.5 & 59.2 & 49.4 & 53.0 & 56.7 & - & - & - & - \\
    CRIS~\cite{wang2022cris}  & ResNet101 & 70.5 & 73.2 & 66.1 & 62.3 & 68.1 & 53.7 & 59.9 & 60.4 & - & - & - & - \\
    LAVT~\cite{yang2022lavt}  & Swin-B & 72.7 & 75.8 & 68.8 & 62.1 & 68.4 & 55.1 & 61.2 & 62.1 & - & - & - & - \\
    ReLA~\cite{liu2023gres}  & Swin-B & \second{73.8} & 76.5 & \second{70.2} & \best{66.0} & \best{71.0} & \second{57.7} & 65.0 & 66.0 & - & - & - & - \\
    X-Decoder~\cite{zou2023generalized} & DaViT-L & - & - & - & - & - & - & 64.6 & - & 22.6 & 17.9 & 21.7 & 16.3 \\
    SEEM~\cite{zou2024segment} & DaViT-L & - & - & - & - & - & - & 65.7 & - & 25.5 & 21.2 & 24.3 & 18.7 \\
    LISA~\cite{lai2024lisa} & LLaVA-7B & \best{74.9} & \best{79.1} & \best{72.3} & \second{65.1} & 
    \second{70.8} & \best{58.1} & \second{67.9} & \best{70.6} & 52.9 & 
54.0 & 47.3 & 48.4 \\
    VISA~\cite{yan2024visa}   & Chat-UniVi-7B & 72.4 & 75.5 & 68.1 & 59.8 & 64.8 & 53.1 & 65.5 & 66.4 & 52.7 & \second{57.8} & - & - \\
    VideoLISA~\cite{bai2024one}    & LLaVA-Phi-3-V & \second{73.8} & 76.6 & 68.8 & 63.4 & 68.8 & 56.2 & \best{68.3} & \second{68.8} & \best{61.4} & \best{67.1} & \best{53.8} & \best{54.4} \\
    \hline
    VRS-HQ    & Chat-UniVi-7B & 73.5 &  \second{77.5} & 69.5 & 61.7 & 67.6 & 54.3 & 66.7& 67.5 & \second{55.2} & 51.8 & \second{51.7} & \second{52.9} \\
% \bottomrule[1.2pt]
\shline
\end{tabular}
}
\vspace{-1mm}
\end{table*}

\subsection{Implementation Details}
% In our experiments, we employed the Multi-modal Large Language Model Chat-UniVi-7B~\cite{jin2024chat} for video reasoning and understanding and SAM2 as the segmentation and tracking model. 
We fine-tuned Chat-UniVi~\cite{jin2024chat} (our chosen MLLM) using LoRA~\cite{hu2021lora} (rank 8), optimizing the mask decoder and MLP projection layer while freezing other parameters. Training used AdamW (learning rate 0.0003, no weight decay) with a WarmupDecayLR scheduler (100-iteration warmup).  We set the fusion coefficient $\alpha$ to 0.1 and sampled two non-keyframes per video.  Trained for 7500 iterations on a hybrid image/video dataset using four A800 GPUs with DeepSpeed~\cite{rasley2020deepspeed} (batch size 1, gradient accumulation 32, total batch size 128), the model employed TDA for video data (token fusion, segmentation, and propagation) and the \TAK token for direct segmentation on image data.

% We chose Chat-UniVi as our MLLM and fine-tuned it using LoRA~\cite{hu2021lora} (rank 8), fully optimizing the mask decoder and MLP projection layer while freezing other parameters.  Training employed AdamW (learning rate 0.0003, no weight decay) with a WarmupDecayLR scheduler (100 iterations warmup).  We set the weighted fusion coefficient $\alpha$ to 0.1 and sampled two non-keyframes per video during training.  The model was trained for 7500 iterations on a hybrid image/video dataset using four NVIDIA A800 GPUs with DeepSpeed~\cite{rasley2020deepspeed} (batch size 1, gradient accumulation 32, total batch size 128).   Video data utilized the TDA strategy for token fusion, segmentation, and propagation across sampled frames.  Image data employed the \TAK token for direct single-frame segmentation.

\subsection{Comparison Results}
\label{Comparison Results}
To showcase the robust pixel-level perception and generalization of VRS-HQ, we conduct evaluations across diverse benchmarks, including ReVOS, RVOS, and image-based referring and reasoning segmentation datasets. 
\vspace{-3mm}

\paragraph{ReVOS Datasets.}
\textcolor{red}{Tab.}~\ref{ReasonVOS} illustrates the performance comparison with previous methods~\cite{botach2022end,wu2022language,lai2024lisa,zhu2023tracking,yan2024visa} on the ReVOS benchmark.  VRS-HQ demonstrates significant improvements over the previous state-of-the-art, VISA~\cite{yan2024visa} across all ten metrics. 
Remarkably, in terms of the $\mathcal{J}$\&$\mathcal{F}$, VRS-HQ-13B surpasses VISA-13B by \textbf{5.9}\% on the referring subset and \textbf{12.5}\% on the reasoning subset, respectively. These gains highlight the effectiveness of our temporal aggregation strategy and the utilization of the \TAK token for keyframe selection and target segmentation, which enhances the model's reasoning capabilities. 
In contrast, VISA relies on LLaMA-VID~\cite{li2025llama} for keyframe selection during inference.  This strategy may neglect intra-frame fine-grained visual details, leading to inaccurate keyframe localization. Furthermore, the robustness score $\mathcal{R}$ exceeds VISA-13B by \textbf{4.4}\%, indicating VRS-HQ’s superior capability in handling negative samples.
% Additionally, our approach significantly surpasses previous methods in hallucination evaluation metrics $\mathcal{R}$, indicating that effectively leveraging temporal tokens for keyframe selection and other aspects assists in mitigating model hallucination.
% the strong video segmentation capabilities of VRS-HQ stem from its temporal dynamic aggregation strategy and its effective utilization of the \TAK token for keyframe selection and target segmentation.

\vspace{-5mm}
\paragraph{RVOS.}
\textcolor{red}{Tab.}~\ref{RefVOS} compares VRS-HQ with state-of-the-art RVOS methods.  On Ref-YouTube-VOS and Ref-DAVIS17, VRS-HQ-13B surpasses VideoLISA~\cite{bai2024one} and VISA-13B~\cite{yan2024visa}, achieving $\mathcal{J}\&\mathcal{F}$ improvements of \textbf{7.3}\% and \textbf{5.6}\%, respectively.  Furthermore, on the motion-intensive MeViS dataset, VRS-HQ-13B yields substantial gains of \textbf{6.2}\%, \textbf{6.1}\% and \textbf{6.4}\% in $\mathcal{J}$, $\mathcal{F}$ and $\mathcal{J}\&\mathcal{F}$, respectively. demonstrating its effective in capturing motion and maintaining temporal coherence.  These improvements, compared to VideoLISA's single-token representation, are attributed to the enhanced inter-frame perception of the TDA and the refined keyframe localization of the TKS.

% \textcolor{red}{Tab.}~\ref{RefVOS} presents a comparison with state-of-the-art RVOS methods. VRS-HQ-7B surpasses previous leading methods, VideoLISA~\cite{bai2024one} and VISA-13B~\cite{yan2024visa}, on the Ref-YouTube-VOS and Ref-DAVIS17 validation sets, achieving $\mathcal{J}\&\mathcal{F}$ improvements of 6.7\% and 5.6\%, respectively.  On the motion-intensive MeViS dataset, VRS-HQ-7B yields substantial gains of 5.8\%/6.1\%/6.1\% across all three metrics, demonstrating its effectiveness in capturing motion transitions and maintaining temporal coherence.  These improvements, compared to VideoLISA's single-token video representation, are attributed to the enhanced inter-frame perception provided by the TDA module and the refined keyframe localization achieved by the TKS module.
% \begin{figure}
%     \centering
%      \vspace{-3mm}
%     \includegraphics[width=0.9\linewidth]{author-kit-CVPR2025-v3.1-latex-/Figures/alpha_fusion_plot.pdf}
%     \vspace{-4mm}
%     \caption{The ablation analysis of the fusion coefficient $\alpha$. 
%     The experimental results suggest that the $\alpha$ value of 0.1 yields the best performance.
% }
%     \label{fusion coefficient}
%     \vspace{-3mm}
% \end{figure}

\begin{table}[htbp]
\centering
\vspace{-2mm}
\caption{Ablation analysis of the fusion coefficient $\alpha$. }
\label{fusion coefficient}
\vspace{-2mm}
\scalebox{0.9}{
\begin{tabular}{c || c c c c c c}
\shline
\rowcolor{gray!35} & \multicolumn{3}{c}{referring} & \multicolumn{3}{c}{reasoning} \\
\cline{2-7}
\rowcolor{gray!35} \multirow{-2}{*}{$\alpha$}& $\mathcal{J}$ & $\mathcal{F}$ & $\mathcal{J}\&\mathcal{F}$ & $\mathcal{J}$ & $\mathcal{F}$ & $\mathcal{J}\&\mathcal{F}$ \\
\shline
    0       & 56.5 & 61.1 & 58.8 & 51.5 & 56.7 & 54.1\\
    0.1     &\best{59.8} & \best{64.5} & \best{62.1} & \best{53.5} & \best{58.7}	& \best{56.1}\\
    0.25    & \second{58.9} & \second{63.8} & \second{61.3} & \second{52.3} & \second{57.8} & \second{55.0}\\
    0.5     & 58.2 & 63.0 & 60.6 & 51.4 & 56.6 & 54.0\\
\shline
\end{tabular}
}
\vspace{-1mm}
\end{table}
\begin{table}[htbp]
\centering
% \vspace{-2mm}
\caption{Ablation analysis of Token-driven Keyframe Selection. $S_{1}$, $S_{2}$, and $S_{3}$ represent CLIP scores, token similarity scores, and occlusion scores, respectively. }
\label{TKS Ablation}
\vspace{-2mm}
\scalebox{0.9}{
\begin{tabular}{c c c || c c c c c c}
\shline
\rowcolor{gray!35} & & & \multicolumn{3}{c}{referring} & \multicolumn{3}{c}{reasoning} \\
% \cline{4-9}
\rowcolor{gray!35} \multirow{-2}{*}{$S_{1}$} & \multirow{-2}{*}{$S_{2}$} & \multirow{-2}{*}{$S_{3}$}& $\mathcal{J}$ & $\mathcal{F}$ & $\mathcal{J}\&\mathcal{F}$ & $\mathcal{J}$ & $\mathcal{F}$ & $\mathcal{J}\&\mathcal{F}$ \\
\shline
{$\surd$} & {$\surd$} & {$\surd$} & \second{59.7} & \second{64.2} & \second{61.9} & \second{52.5} & \second{57.8} & \second{55.2} \\
% {$\surd$} & & & 57.8 & 62.4 & 60.1 & 50.9 & 56.2 & 53.5 \\
%  & {$\surd$} & & 58.4 & 63.0 & 60.7 & 51.3 & 56.4 & 53.8 \\
%  & & {$\surd$} & 59.9 & 64.6 & 62.2 & 53.4 & 58.5 & 55.9 \\
{$\surd$} & {$\surd$} & &57.8 & 62.4 & 60.1 & 50.9 & 56.2 & 53.6 \\
{$\surd$} &  & {$\surd$} &59.6 & 64.1 & 61.8 & \second{52.5} & 57.7 & 55.1 \\  
 & {$\surd$} & {$\surd$} & \best{59.8} & \best{64.5} & \best{62.1} & \best{53.5} & \best{58.7} & \best{56.1}\\
 
\shline
\end{tabular}}
\vspace{-3mm}
\end{table}

\vspace{-5mm}
\paragraph{RIS.}
Our method seamlessly extends to referring image segmentation (RIS) by treating images as single-frame videos.  As shown in \textcolor{red}{Tab.}~\ref{RefCOCO}, VRS-HQ consistently outperforms VISA on three RIS benchmarks, achieving performance comparable to LISA~\cite{lai2024lisa} and VideoLISA~\cite{bai2024one}. On ReasonSeg, VRS-HQ performs competitively, ranking merely below VideoLISA, demonstrating strong generalization.  The marginally lower performance on image datasets compared to video datasets likely stems from two factors: (i) our model's training emphasis on video data (unlike LISA, which is pre-trained solely on images); and (ii) our training methodology encourages reliance on SAM2's memory mechanism for multi-frame processing, which may be less effective for single-image segmentation compared to VideoLISA's single-token approach.

\subsection{Ablation Studies}
\label{Ablation Study}
% To validate each module's contribution, we conduct extensive ablation studies on the ReVOS dataset, analyzing the fusion coefficient $\alpha$ within the TDA module, the impact of different score combinations in the TKS module, the mask decoding and propagation strategy, and the frame sampling strategy during inference.  
% role of SAM2’s memory storage mechanism during training and inference. 
% We initially examine the key components of VRS-HQ, specifically exploring the fusion coefficient $\alpha$ within the TDA module and the TKS module. Additionally, we investigate the role of SAM2’s memory storage mechanism in both training and inference. 
%, as well as the effect of different video frame sampling strategies during inference.
% \vspace{-5mm}
\paragraph{Fusion Coefficient Ablation.}
The ablation analysis of the fusion coefficient $\alpha$ in the TDA is illustrated in \textcolor{red}{Tab.}~\ref{fusion coefficient}. 
An $\alpha$  of 0.1 yields optimal performance.  Setting $\alpha$ to 0 leads to a sharp drop in the model's performance as \TAK fails to capture fine-grained intra-frame details, and the \SEG token cannot be jointly optimized with \TAK through backpropagation (\eg, \textbf{62.1}\%$\to$\textbf{58.8}\% $\mathcal{J}$\&$\mathcal{F}$ on the referring subset, \textbf{56.1}\%$\to$\textbf{54.1}\% $\mathcal{J}$\&$\mathcal{F}$ on the reasoning subset). 
As $\alpha$ increases beyond 0.1, the metrics decline slightly, likely due to excessive frame-level noise being introduced into the temporal token at a higher fusion coefficient (\eg, \textbf{62.1}\%$\to$\textbf{60.6}\% $\mathcal{J}$\&$\mathcal{F}$  on the referring subset, \textbf{56.1}\%$\to$\textbf{54.0}\% $\mathcal{J}$\&$\mathcal{F}$ on the reasoning subset). 

\vspace{-3mm}

\begin{figure*}
    \centering
    \vspace{-2mm}
    \includegraphics[width=0.98\linewidth]{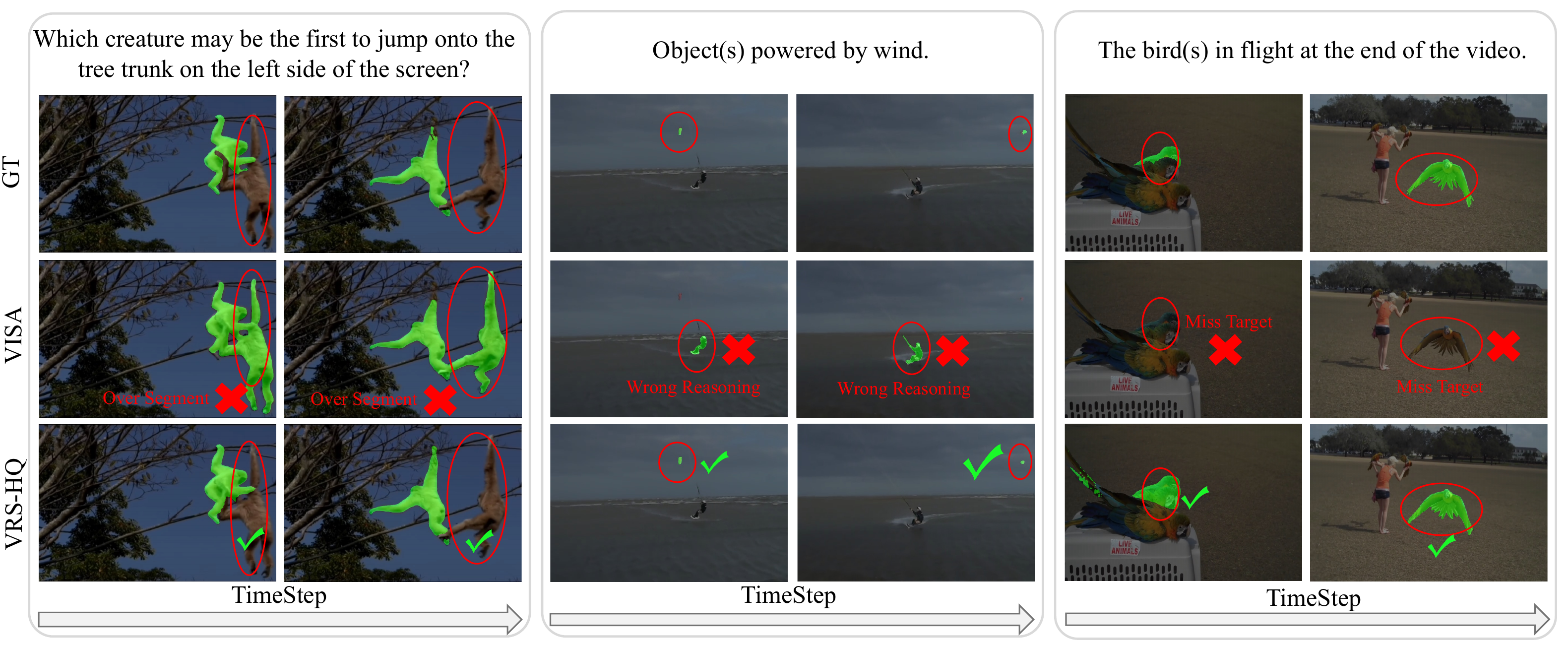}
    \vspace{-2mm}
    \caption{Segmentation map comparison of VISA and VRS-HQ on the ReVOS benchmark (\textsection\ref{results}). Results across three scenarios demonstrate that VRS-HQ excels in reasoning complex spatial and temporal relationships, delivering enhanced segmentation performance. 
    }
    \vspace{-3mm}
    \label{result visualization}
    % \vspace{-2mm}
\end{figure*}

\begin{figure}
    \centering
    % \vspace{-3mm}
    \includegraphics[width=1.02\linewidth]{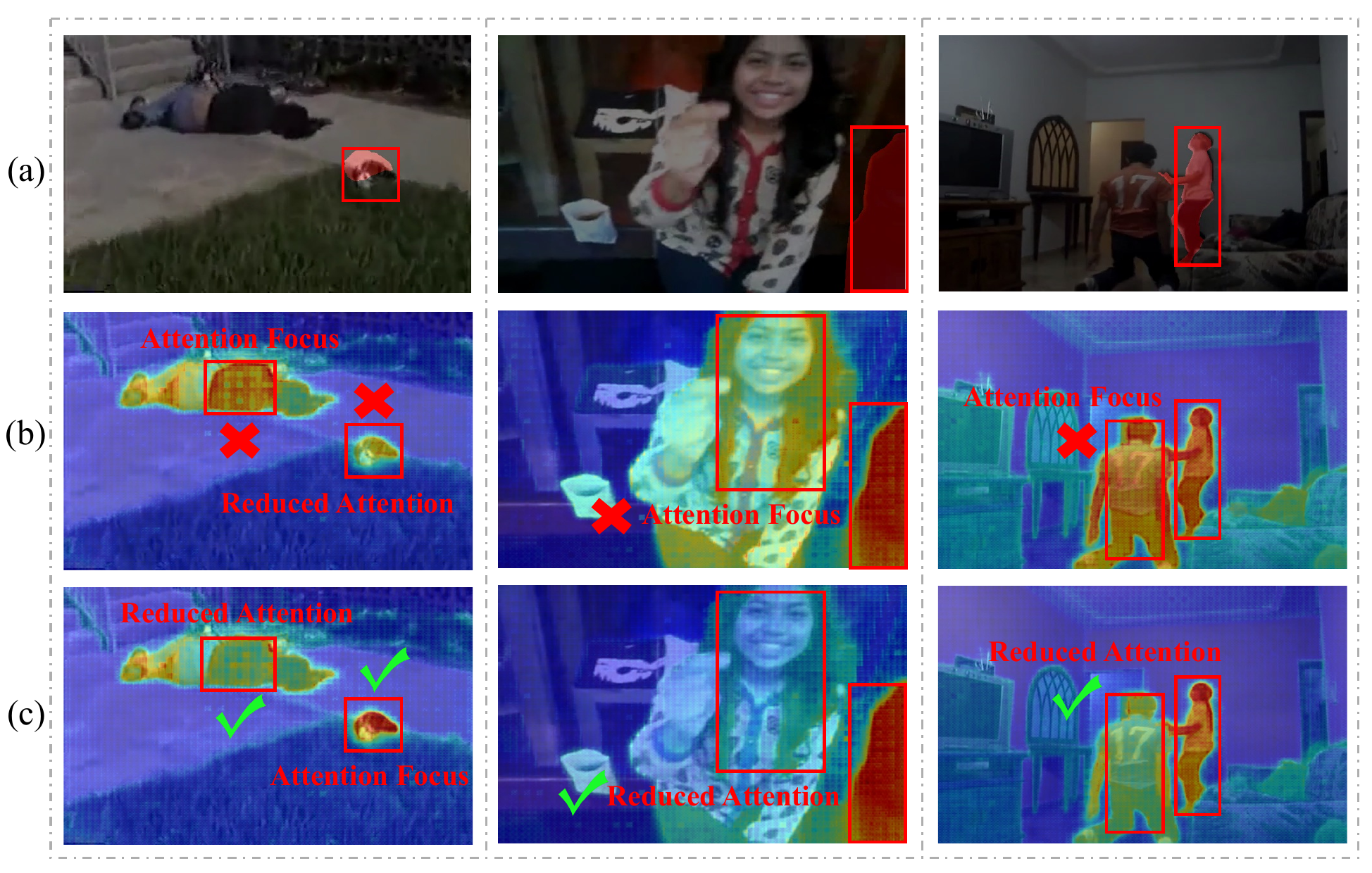}
    \vspace{-5mm}
    \caption{Visualization of feature maps (\textsection\ref{feature}). From top to bottom are:  (a) Ground truth masks. b) Keyframe mask embeddings generated by the \TAK token \textit{before} TDA. (c) Keyframe mask embeddings generated by the \TAK token \textit{after} TDA.
}
\vspace{-3mm}
    \label{feature visualization}
    % \vspace{-1mm}
\end{figure}

\paragraph{Token-driven Keyframe Selection.}   % 对于这个表格的描述？
\textcolor{red}{Tab.}~\ref{TKS Ablation} analyzes the impact of different score combinations for keyframe selection within the TKS.  Results indicate that the occlusion score is the most influential of the three considered, improving $\mathcal{J}\&\mathcal{F}$ by \textbf{1.8}\% and \textbf{1.6}\% on the referring and reasoning subsets, respectively (first two rows).  This improvement stems from the occlusion score's ability to reflect the target object's presence within each frame. The model achieves its best performance when keyframe selection is based on a combination of token similarity and the occlusion score.  However, incorporating the CLIP score (rows 1 and 4) slightly degrades performance (by \textbf{0.2}\% and \textbf{0.9}\% $\mathcal{J}\&\mathcal{F}$ on the referring and reasoning subsets, respectively), likely due to CLIP's limited spatiotemporal understanding, which hinders accurate keyframe selection.

% Within the TKS module, we explore the impact of various score combinations on keyframe selection, as depicted in \textcolor{red}{Tab.}~\ref{TKS Ablation}.
% Experimental results indicate that the occlusion score is the most influential among the three, advancing an improvement of \textbf{1.8}\% and \textbf{1.5}\% $\mathcal{J}$\&$\mathcal{F}$ on the referring and reasoning subset separately, as it reflects the likelihood of the target's presence in each frame.
% The model performs best when keyframe selection is based on a combination of token similarity and occlusion score. 
% From the results in the first and last row, however, it can be seen that adding the \texttt{CLIP} score leads to a slight performance drop of \textbf{0.2}\% and \textbf{1.0}\% $\mathcal{J}$\&$\mathcal{F}$ for the referring and reasoning subsets, likely due to \texttt{CLIP}’s limited fine-grained spatial and temporal understanding, hindering keyframe selection accuracy.
% However, adding the CLIP score slightly lowers performance, likely due to CLIP’s limited fine-grained spatial and temporal understanding, hindering keyframe selection accuracy.
% 因为论文空间有限，这个实验意义不大，先不放上来了，后续可以放在补充材料里
% \paragraph{Sampling Strategy During Inference}   % 看情况是否去掉
\begin{table}[tbp]
\centering
\vspace{-2mm}
\caption{Ablation analysis of the mask decoding and propagation strategy. MT/MI: Multi-frame Training/Inference. ST/SI: Single-frame Training/Inference.}
\label{SAM2 Ablation}
\vspace{-2mm}
\scalebox{0.85}{
\begin{tabular}{c || c c c c c c}
\shline
\rowcolor{gray!35} & \multicolumn{3}{c}{referring} & \multicolumn{3}{c}{reasoning} \\
\cline{2-7}
\rowcolor{gray!35} \multirow{-2}{*}{Strategy}& $\mathcal{J}$ & $\mathcal{F}$ & $\mathcal{J}\&\mathcal{F}$ & $\mathcal{J}$ & $\mathcal{F}$ & $\mathcal{J}\&\mathcal{F}$ \\
\shline
SAM+ST+SI & 48.3 & 52.4 & 50.3 & 42.9 & 47.4 & 45.2 \\
SAM2+ST+SI & \second{55.9} & 59.8 & 57.8 & \second{49.3} & \second{54.0} & \second{51.6}\\ 
SAM2+ST+MI & 55.8 & \second{60.8} & \second{58.3} & 46.5 & 52.5 & 49.5 \\
SAM2+MT+SI & 54.1 & 57.4 & 55.8 & 48.2 & 52.2 & 50.2\\
SAM2+MT+MI & \best{59.8} & \best{64.5} & \best{62.1} & \best{53.5} & \best{58.7} & \best{56.1}\\
\shline
\end{tabular}
}
\vspace{-1mm}
\end{table}
\begin{table}[tbp]
\centering
% \vspace{-2mm}
\caption{Ablation analysis of sampling strategy.}
% during inference
\vspace{-2mm}
\label{Sampling Strategy Ablation}
\scalebox{0.85}{
\begin{tabular}{c || w{c}{0.64cm} w{c}{0.64cm} w{c}{0.64cm} w{c}{0.64cm} w{c}{0.64cm} w{c}{0.64cm}}
\shline
\rowcolor{gray!35}& \multicolumn{3}{c}{referring} & \multicolumn{3}{c}{reasoning} \\
% \cline{2-7}
\rowcolor{gray!35}\multirow{-2}{*}{Sampling strategy}& $\mathcal{J}$ & $\mathcal{F}$ & $\mathcal{J}\&\mathcal{F}$ & $\mathcal{J}$ & $\mathcal{F}$ & $\mathcal{J}\&\mathcal{F}$ \\
\shline
Random Sampling& 59.0& 63.6& 61.3& 53.2 & \second{58.4}& \second{55.8}\\
Uniform Sampling & \second{59.3}& \second{63.9}& \second{61.6}& \second{53.3}& \second{58.4}& \second{55.8}\\
CLIP Sampling & \best{59.8} & \best{64.5} & \best{62.1} & \best{53.5}& \best{58.7}& \best{56.1}\\ 
\shline
% \bottomrule[1.2pt]
\end{tabular}
}
\vspace{-3mm}
\end{table}
\vspace{-3mm}
\paragraph{Mask Decoding and Propagation.}
\textcolor{red}{Tab.}~\ref{SAM2 Ablation} analyzes the impact of different mask decoding and propagation strategies on segmentation performance.  Using only SAM and TDA for single-frame segmentation (first row) already surpasses VISA-13B by \textbf{0.9}\%/\textbf{0.7}\%/\textbf{0.9}\% on the reasoning subset.  Comparing the first two rows reveals the substantial advantage of SAM2 over SAM for single-frame fine-tuning and inference.  SAM2 achieves improvements of \textbf{7.6}\%/\textbf{7.4}\%/\textbf{7.5}\%/\textbf{6.4}\%/\textbf{6.6}\%/\textbf{6.4}\%, demonstrating its superior robustness in segmentation and tracking. Furthermore, incorporating multi-frame propagation during both training and inference (last row) yields the best overall results, effectively leveraging SAM2's memory mechanism for keyframe mask storage. This approach leads to further improvements in $\mathcal{J}\&\mathcal{F}$ of \textbf{3.8}\% and \textbf{4.5}\% compared to the other strategies.
% According to \textcolor{red}{Tab.}~\ref{SAM2 Ablation}, we investigate the impact of various mask decoding and propagation strategies on the segmentation performance.
% In the first row, it is evident that our method, using only the SAM and TDA for single-frame segmentation, surpasses VISA-13B by \textbf{0.9}\%/\textbf{0.7}\%/\textbf{0.9}\% in the reasoning subset. 
% From the first two rows of the table, we can observe that while using single-frame fine-tuning and inference, using SAM2 outperforms SAM by \textbf{7.6}\%/\textbf{7.4}\%/\textbf{7.5}\%/\textbf{6.4}\%/\textbf{6.6}\%/\textbf{6.4}\% across six metrics, demonstrating its superior robustness in segmentation and tracking.
% Moreover, the results in the last row indicate that adopting multi-frame propagation during both training and inference yields the best performance, effectively utilizing SAM2's memory mechanism for keyframe mask storage, achieving an increase of \textbf{3.8}\% and \textbf{4.5}\% $\mathcal{J}$\&$\mathcal{F}$ compared to other strategies. 
\vspace{-6mm}
\paragraph{Inference-Time Sampling Strategy.}
\textcolor{red}{Tab.}~\ref{Sampling Strategy Ablation} analyzes the impact of different frame sampling strategies during inference.  
``Random" denotes randomly selected frames, while ``Uniform" extracts frames at equal intervals.
CLIP sampling slightly outperforms uniform sampling, improving $\mathcal{J}\&\mathcal{F}$ by \textbf{0.5}\%/\textbf{0.3}\% on the referring and reasoning subsets, respectively.  This modest gain is likely due to CLIP's ability in selecting frames that are semantically aligned with the referring expression, which in turn facilitates the reasoning process of the MLLM. 
% \paragraph{Sampling Strategy During Inference.}
% As shown in \textcolor{red}{Tab.}~\ref{Sampling Strategy Ablation}, we apply different  frame sampling strategies during inference. The \texttt{CLIP} sampling approach performs slightly better, surpassing uniform sampling by \textbf{0.5}\%/\textbf{0.3}\% in the $\mathcal{J}$\&$\mathcal{F}$ for the referring and reasoning subset. 
% % , while being \textbf{0.1}\% lower in the reasoning subset. 
% This discrepancy may be attributed to \texttt{CLIP}’s ability to coarsely extract frames with the highest semantic similarity to the expression, providing a degree of optimization for MLLM’s reasoning.

\subsection{Visualization Analysis}   
% \vspace{-2mm}
\paragraph{Segmentation Map Comparison.}
\label{results}
\textcolor{red}{Fig.}~\ref{result visualization} presents the qualitative comparison of VRS-HQ and VISA on the ReVOS dataset across three different scenarios. 
In the left case, VISA displays ambiguity in spatial positioning and neglects critical references such as ``the first", revealing its limited semantic understanding. 
% In the leftmost case, VISA struggles with intricate spatial concepts, displaying ambiguity in spatial positioning and neglecting critical references such as "the first.", underscoring its limited semantic understanding. 
In contrast, VRS-HQ demonstrates a strong ability to perceive spatiotemporal relationships.
For the middle example, VISA mistakenly segments the person flying the kite, while VRS-HQ correctly identifies the kite, reflecting the stronger capacity to apply world knowledge and better handling of small-object segmentation.
% The middle example accesses the model’s capacity to apply world knowledge in video reasoning. While VISA mistakenly segments the person flying a kite, VRS-HQ correctly identifies the kite, reflecting a deeper semantic understanding and better handling of small-object segmentation.
% reflecting a more nuanced semantic understanding and the ability to handle small-object segmentation. 
% This demonstrates VRS-HQ’s superior semantic understanding and ability to handle small-object segmentation.
The right column showcases VRS-HQ's robust temporal reasoning capabilities, driven by our temporal dynamic aggregation and adaptive keyframe selection. Conversely, VISA misses the target due to incorrect keyframe selection during inference.
\vspace{-5mm}

% \paragraph{Feature Visualization.} To better understand the impact of the TDA module, we visualize the mask embeddings produced by the mask decoder. These embeddings are generated using both the fused and unfused TAK tokens in conjunction with the keyframe features.  As illustrated in Figure~\ref{feature visualization}, we employ Principal Component Analysis (PCA)~\cite{wold1987principal} to reduce the dimensionality of the mask embeddings to a single channel. This single-channel representation is then blended with the original image using a scaling factor of 0.5. This blending technique effectively highlights the regions of high attention within the image.  Prior to the application of the TDA module, the visualization reveals that the TAK token frequently focuses on elements that are not the intended targets.  For instance, in the first example, the focus is on the falling person; in the second example, the focus is on the woman; and in the third example, the focus is on the dancing man in the center of the frame.  In contrast, after the application of the TDA module, the similarity-weighted fusion of the frame-level SEG tokens into the TAK token significantly enriches the positional and semantic information contained within the token. This enrichment allows the model to more effectively concentrate on the actual target objects within the scene.

\paragraph{Feature Visualization.}
\label{feature}
To understand the TDA's impact, we visualize mask embeddings generated by the mask decoder using both the fused and unfused \TAK tokens with keyframe features.  As illustrated in \textcolor{red}{Fig.}~\ref{feature visualization}, we employ Principal Component Analysis (PCA)~\cite{wold1987principal} to reduce the dimensionality of the mask embeddings to a single channel. This single-channel representation is then blended with the original image to highlight high-attention regions.  Before TDA fusion, the \TAK token often focuses on non-target elements (e.g., the falling person in the first example, the woman in the second, and the central dancing man in the third).  
In contrast, the similarity-weighted fusion of the frame-level SEG tokens into the \TAK token significantly enriches the positional and semantic information contained within the token, enabling the model to more effectively concentrate on the correlated target objects.

% To comprehensively assess the impact of the TDA module, we interact the fused and unfused \TAK token with keyframe features using the mask decoder and visualize the resulting mask embeddings. 
% As depicted in \textcolor{red}{Fig.}~\ref{feature visualization}, we apply the Principal Component Analysis (PCA)~\cite{wold1987principal} to distill the mask embedding to a single channel, blending it with the original image at a scaling factor of 0.5, where high-attention regions are highlighted. The visualization reveals that the \TAK token prior to the TDA module tends to focus on non-targets, \eg the falling person in the first case, the woman in the middle case, and the dancing man in the center of the right example. In contrast, the similarity-weighted fusion of frame-level \SEG tokens into the \TAK token enriches positional and semantic information, allowing the model to better concentrate on the actual targets.

\section{Conclusion}
We present VRS-HQ, a novel approach that leverages the temporal reasoning of MLLM and the robust tracking of SAM2 for high-quality Video Reasoning Segmentation. 
% The core idea of our approach is to enhance the model’s perception and extraction of intra-frame spatial features and inter-frame temporal relations through Language-driven Temporal Dynamic Aggregation. 
% The core idea of our approach is to design the frame and temporal-level special tokens, facilitating temporal collaboration and keyframe selection. 
% To be specific, the MLLM is prompted to generate multi-level special tokens: \SEG and \TAK, incorporating intra-frame spatial features and inter-frame temporal relations respectively. Then the extracted \SEG and \TAK embeddings are sent to the Temporal Dynamic Aggregation module to integrate the information of targets from each frame into the video-level token \TAK. 
% Subsequently, SAM2 utilizes the \TAK token to perform keyframe segmentation and propagates the mask to the remaining frames. 
% Our approach introduces temporal-level \TAK token and frame-level \SEG tokens to facilitate temporal collaboration and keyframe selection. The MLLM generates multi-level tokens: \SEG for intra-frame spatial features and \TAK for inter-frame temporal relations. 
Our method utilizes a temporal-level \TAK token and frame-level \SEG tokens to capture temporal relations and spatial features, respectively. These tokens are integrated using the Temporal Dynamic Aggregation, with SAM2 utilizing the \TAK token for keyframe segmentation and mask propagation. Moreover, we introduce a Token-driven Keyframe Selection that uses the \TAK token to generate occlusion scores for robust keyframe selection. Extensive experiments confirm that VRS-HQ achieves state-of-the-art performance on various benchmarks, demonstrating strong capabilities in handling Video Reasoning Segmentation.
% and offering significant insights for real-world applications.
% \section*{A. Introduction}
\section{Appendix}
This supplementary material provides additional details and analysis of VRS-HQ, expanding on the content presented in the main paper. 
We begin by evaluating the impact of various training datasets on segmentation performance (\textsection\hyperref[datasets ablation]{A}). 
Next, we present more detailed implementation information to facilitate reproducibility (\textsection\hyperref[implementation details]{B}). 
We then elaborate on the specific method of utilizing SAM2~\cite{ravi2024sam} for mask decoding and propagation (\textsection\hyperref[sam2 details]{C}).
Subsequently, we show some failure cases with analysis to offer a more comprehensive understanding of VRS-HQ's limitations (\textsection\hyperref[failure case analysis]{D}).
Additionally, we present more qualitative comparisons against VISA, highlighting the strengths of our proposed method (\textsection\hyperref[qualitative comparison]{E}). 
Finally, we visualize the reasoning segmentation results of VRS-HQ on in-the-wild video datasets, demonstrating its strong generalization capabilities (\textsection\hyperref[in-the-wild]{F}).

\section*{A. Datasets Ablation}
\label{datasets ablation}
As illustrated in \textcolor{red}{Tab.}~\ref{Dataset Ablation}, fine-tuning with the full datasets yields the best performance while excluding the image segmentation dataset, VideoQA dataset~\cite{maaz2023video}, or ReVOS dataset~\cite{yan2024visa} individually results in varying degrees of metric degradation.
Notably, removing the VideoQA dataset minimally impacts the model’s performance, with a decline of \textbf{0.9}\% in $\mathcal{J}\&\mathcal{F}$ on both the referring and reasoning subsets, as its primary role is to support the MLLM’s video comprehension rather than directly contributing to the segmentation process.
In contrast, excluding the ReVOS dataset leads to a noticeable drop of \textbf{4.4}\% and \textbf{7.6}\% in $\mathcal{J}\&\mathcal{F}$, highlighting its pivotal role in enhancing the model’s reasoning segmentation performance in challenging scenarios.
\begin{table}[!htbp]
\centering
% \vspace{-2mm}
\caption{Ablation study on the impact of training datasets.}
\label{Dataset Ablation}
\vspace{-3mm}
\scalebox{0.85}{
\begin{tabular}{c || c c c c c c}
\shline
\rowcolor{gray!35} & \multicolumn{3}{c}{referring} & \multicolumn{3}{c}{reasoning} \\
\cline{2-7}
\rowcolor{gray!35} \multirow{-2}{*}{Datasets}& $\mathcal{J}$ & $\mathcal{F}$ & $\mathcal{J}\&\mathcal{F}$ & $\mathcal{J}$ & $\mathcal{F}$ & $\mathcal{J}\&\mathcal{F}$ \\
\shline
Joint & \best{59.8} & \best{64.5} & \best{62.1} & \best{53.5} & \best{58.7} & \best{56.1} \\
w/o ImageSeg  & 58.5 & 63.2 & 60.8 & 51.0 & 56.3 & 53.6 \\ 
w/o VideoQA & \second{58.7} & \second{63.7} & \second{61.2} & 
\second{52.4} & \second{58.0} & \second{55.2} \\
w/o ReVOS & 55.3 & 60.1 & 57.7 & 45.3 & 51.6 & 48.5 \\
% SAM2+MT+MI & \best{59.8} & \best{64.5} & \best{62.1} & \best{53.5} & \best{58.7} & \best{56.1}\\
\shline
\end{tabular}
}
% \vspace{-1mm}
\end{table}

\section*{B. Additional Implementation Details}
\label{implementation details}
Due to space constraints of the main document, additional implementation details are provided here. 
During training, we use varying sampling ratios for different datasets (\textit{cf.} \textcolor{red}{Tab.}~\ref{dataset sampling}).
For video segmentation datasets, 8-12 frames are uniformly sampled at fixed intervals per video, and up to three object categories are selected per image or video. 
During inference, we utilize CLIP-336~\cite{radford2021learning} for global sampling, selecting up to 12 frames per video. 
% The precise number of frames is dynamically adjusted according to the total frame count to ensure optimal coverage.
% The exact number of sampled frames is adjusted based on the total frame count of the video to ensure optimal coverage.
Input images are resized to $224\times{224}$ before being input to Chat-UniVi~\cite{jin2024chat}. Data passed to SAM2 is augmented as described in~\cite{kirillov2023segment} and resized to $1024\times{1024}$. 
Moreover, LoRA~\cite{hu2021lora} is applied with a scaling factor of 16 and a dropout rate of 0.05 across all query and value projection layers within the MLLM,  
enabling efficient fine-tuning.
% to improve flexibility and reduce overfitting.
% the scaling factor of LoRA~\cite{hu2021lora} is set to 16, while the dropout rate is configured at 0.1. These LoRA modules are integrated into all query and value projection layers of the MLLM.
\setlength{\tabcolsep}{3.5pt}
\begin{table}[!htbp]
\centering
\vspace{-2mm}
\caption{Datasets sampling ratio during training.}
\label{dataset sampling}
\vspace{-3mm}
\scalebox{0.85}{
\begin{tabular}{l | c c c c c c}
\shline
Dataset & SemSeg & RIS & ImageQA & ReaSeg & VideoQA & VideoSeg \\
\shline
Ratios & 9/32 & 3/32 & 3/32 & 1/32 & 1/8 & 3/8 \\
\shline
\end{tabular}
}
\vspace{-2mm}
\end{table}

\section*{C. More Details of SAM2}
\label{sam2 details}
\begin{figure}
    \centering
    % \vspace{-2mm}
    \includegraphics[width=1\linewidth]{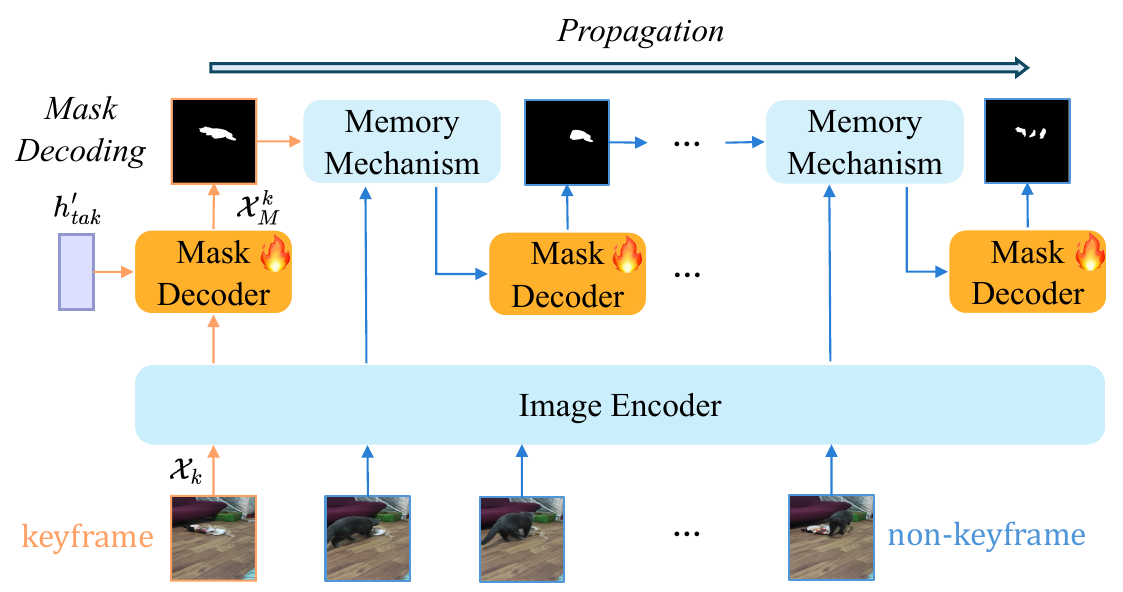}
    \vspace{-2mm}
    \caption{Details of  SAM2 for mask decoding and propagation. All the video frames are input into the image encoder for feature extraction. The feature embeddings of the keyframe interact with $h'_{tak}$ through the mask decoder for mask generation and then propagate it to the remaining video frames via the memory mechanism. 
   %  (b) The memory mechanism of SAM2 comprises a memory encoder for previous mask encoding, stores them in a memory bank, and utilizes memory attention to facilitate interaction between the input features and the stored mask features, enabling cross-frame propagation. 
   }
    \vspace{-1mm}
    \label{SAM2}
    % \vspace{-3mm}
\end{figure}

As depicted in \textcolor{red}{Fig.}~\ref{SAM2}, we provide detailed insights into the process of mask decoding and propagation using SAM2~\cite{ravi2024sam}.
Specifically, all input video frames are processed through the image encoder to extract multi-scale visual features. 
Subsequently, the fused temporal embedding $h'_{tak}$ interacts with the keyframe features in the mask decoder to generate the segmentation mask and perform video-level propagation. 
The prediction is then encoded by the memory encoder and stored in the memory bank, which maintains a FIFO queue of memories from recent frames. % to manage previously stored segmentation masks. 
Feature embeddings from subsequent non-keyframes attend to these stored mask features through memory attention and utilize the mask decoder to generate corresponding masks, enabling inter-frame propagation.

\begin{figure*}[t!]
    \centering
    % \vspace{-2mm}
    \includegraphics[width=0.98\linewidth]{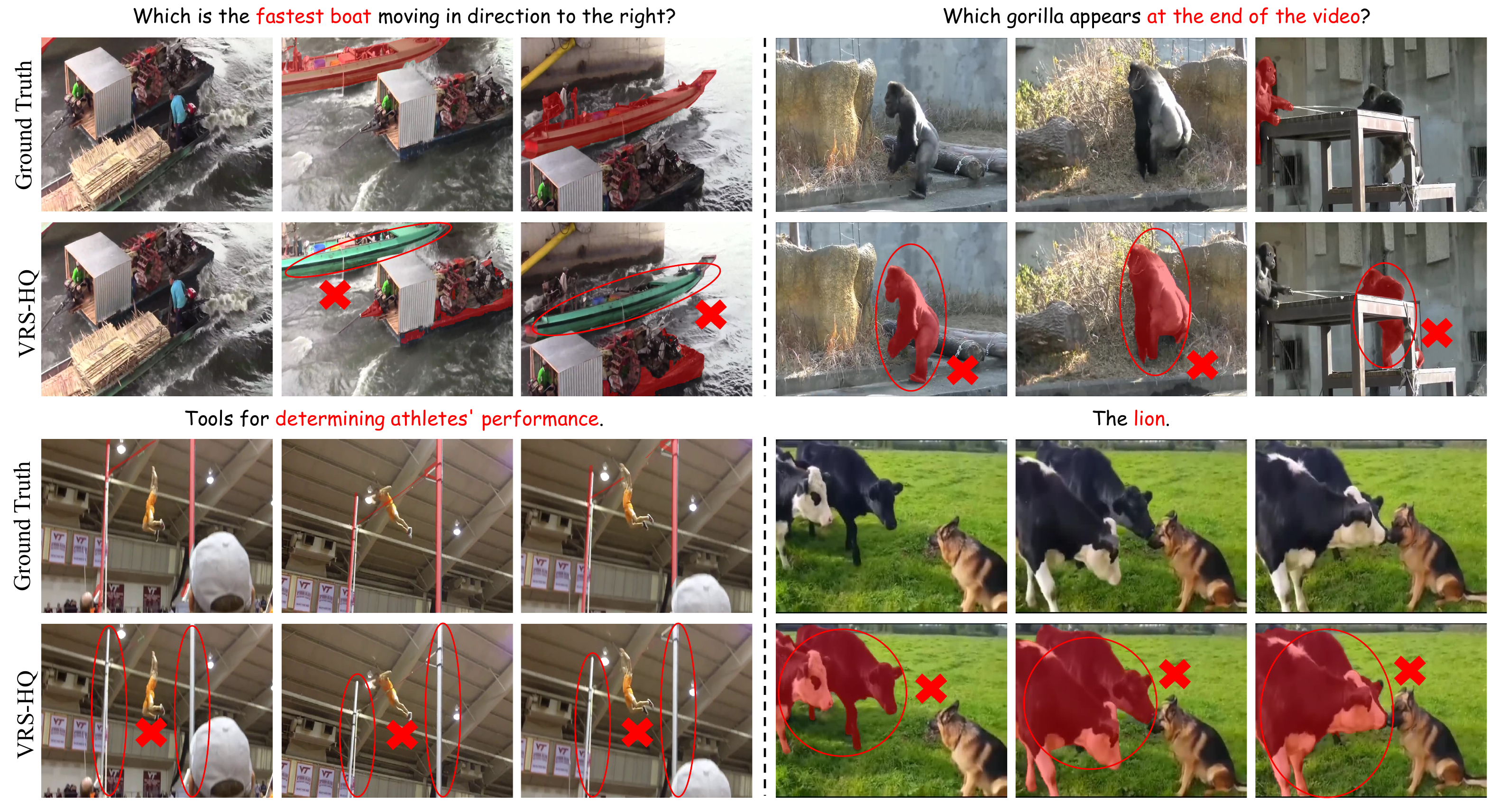}
    \vspace{-1mm}
    \caption{Visualization of failure cases for VRS-HQ. These examples illustrate the model's limitations in scenarios requiring complex world knowledge and temporal reasoning, as well as challenges in processing negative samples. 
    }
    % \vspace{-5mm}
    \label{failure cases}
    % \vspace{-2mm}
\end{figure*}

\section*{D. Failure Case Analysis}   % 可能一段原因还不太够，还需要再增加一些
\label{failure case analysis}
% As shown in Fig.~\ref{failure cases}, we provide and analyze several failure cases to gain a deeper understanding of VRS-HQ's performance limitations.
\textcolor{red}{Fig.}~\ref{failure cases} presents a detailed analysis of several failure cases, offering a deeper understanding of the limitations of VRS-HQ. The \textbf{top row} highlights two specific challenges. \textbf{First}, VRS-HQ struggles with keyframe localization when presented with queries based on motion, such as identifying the fastest-moving boat within a video sequence.  This suggests a potential weakness in analyzing and interpreting dynamic visual information. \textbf{Second}, the model exhibits difficulty segmenting targets with minimal temporal presence, as exemplified by the gorilla visible only in the last two frames of the video. This points to a possible limitation in effectively capturing and utilizing short-duration visual cues.
The \textbf{bottom row} reveals further limitations. VRS-HQ demonstrates a lack of comprehension when faced with nuanced or implicitly phrased prompts, such as recognizing a ``high bar" within the context of gymnastics performance evaluation. This suggests a need for improved understanding of complex semantic relationships within video content.  Furthermore, the model occasionally exhibits hallucinatory behavior, generating segmentations for non-existent objects, particularly when dealing with empty targets or scenes where the requested object is absent.

% The top row highlights challenges in keyframe localization for motion-based queries (e.g., identifying the fastest-moving boat) and in segmenting objects with limited visibility (e.g., the gorilla appearing only in the final two frames). This suggests weaknesses in processing high-dynamic visual information and capturing short-duration cues. The bottom row demonstrates difficulties in comprehending nuanced prompts (e.g., recognizing a ``high bar" in gymnastics) and occasional hallucinatory behavior with empty targets.

We hypothesize that several strategies could mitigate these limitations.  Improving the video comprehension capabilities of the Multimodal Large Language Model (MLLM) could enhance the ability to interpret complex scenes and queries.  Enabling the model to process a larger number of sampled frames simultaneously might improve its sensitivity to subtle temporal changes and short-duration events.  Finally, designing specialized tokens specifically for representing empty masks could address the observed hallucinations in such scenarios.  We leave a thorough investigation of these potential improvements to future research.

% We can observe from the left case that VRS-HQ struggles to identify the target when the prompt is overly complex, such as recognizing a ``high bar" used in gymnastics performance evaluation, which likely stems from the MLLM’s restricted video comprehension capabilities. 
% In the right case, VRS-HQ fails to accurately locate the keyframe and identify the fastest-moving boat, indicating that understanding expressions related to intricate motion and extracting global video features remain challenging areas for VRS-HQ. 
% We plan to further investigate these limitations in our future work.

\section*{E. More Qualitative Comparison}
\label{qualitative comparison}
\begin{figure*}[t!]
    \centering
    \vspace{-2mm}
    \includegraphics[width=1\linewidth]{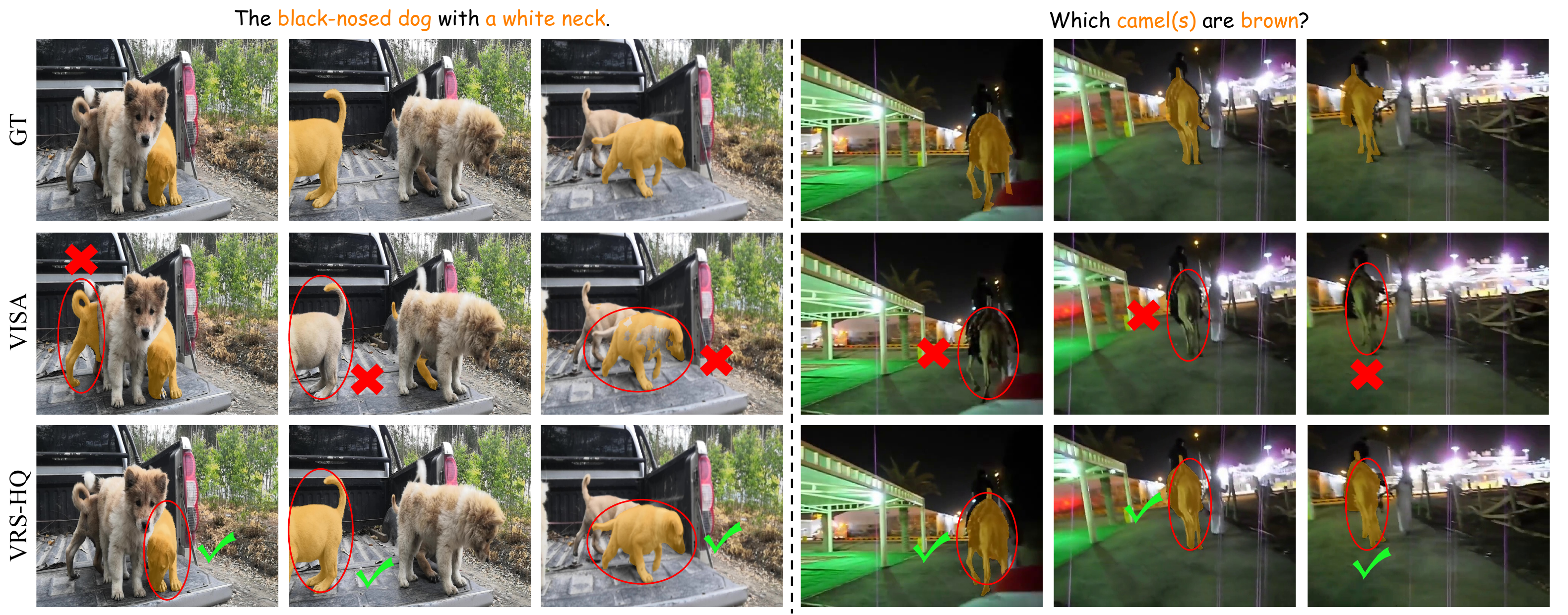}
    \vspace{-6mm}
    \caption{
    Qualitative comparison of VRS-HQ and VISA in explicit language-based referring scenarios on the ReVOS benchmark. 
    % Segmentation map comparison of VISA and VRS-HQ on the ReVOS benchmark.  
    }
    % \vspace{-1mm}
    \label{result visualization1}
    % \vspace{-2mm}
\end{figure*}

\begin{figure*}[t!]
    \centering
    % \vspace{-3mm}
    \includegraphics[width=1\linewidth]{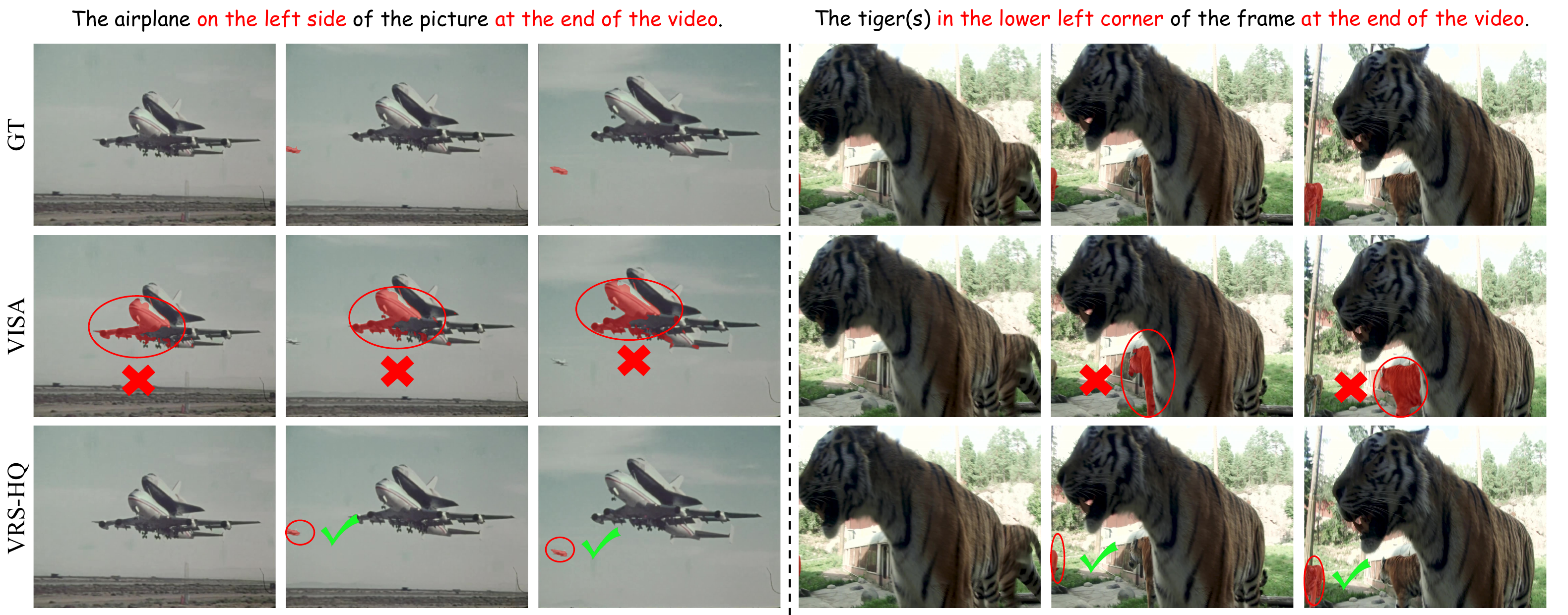}
    \vspace{-5mm}
    \caption{Qualitative comparison of VRS-HQ and VISA in scenarios incorporating complex temporal dynamics on the ReVOS benchmark. 
    }
    % \vspace{-2mm}
    \label{result visualization2}
    % \vspace{-4mm}
\end{figure*}

\begin{figure*}[t!]
    \centering
    \vspace{-1mm}
    \includegraphics[width=1\linewidth]{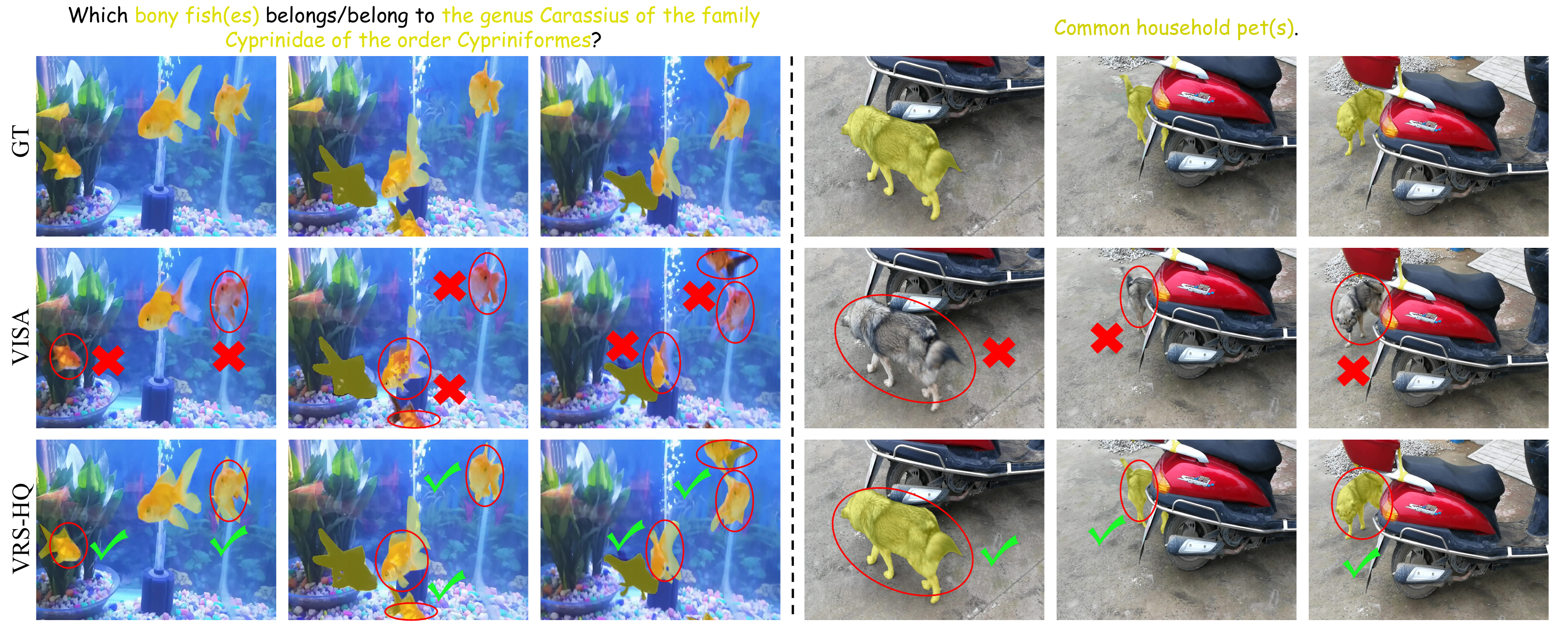}
    \vspace{-5mm}
    \caption{
    Qualitative comparison of VRS-HQ and VISA in reasoning scenarios that require world knowledge on the ReVOS benchmark.
    }
    \vspace{-3mm}
    \label{result visualization3}
    \vspace{-2mm}
\end{figure*}

In addition to the visual comparisons presented in the main document, we provide further comparisons across more diverse settings in \textcolor{red}{Fig.}~\ref{result visualization1}-\ref{result visualization3} to demonstrate the model's reasoning and segmentation capabilities.
As illustrated in \textcolor{red}{Fig.}~\ref{result visualization1}, VISA demonstrates reduced sensitivity to color-related expressions (e.g., ``white" and ``brown") when provided with explicit textual instructions. Furthermore, the example on the left demonstrates VISA's tendency to misidentify visually similar objects with complex spatial variations.  In contrast, VRS-HQ effectively aggregates temporal information, capturing inter-frame motion dynamics and leading to improved segmentation accuracy.
% Moreover, the left case reveals VISA’s tendency to misidentify multiple visually similar objects with intricate spatial variations.  
% Conversely, VRS-HQ effectively aggregates temporal dynamics, capturing inter-frame motion dynamics and enhancing segmentation accuracy.

\textcolor{red}{Fig.}~\ref{result visualization2} highlights the robust segmentation and reasoning capabilities of VRS-HQ in scenarios with complex temporal dynamics. 
In the left example, VISA struggles to precisely detect the airplane appearing on the left at the end of the video. 
Similarly, in the right case, VISA misclassifies the tiger emerging in the lower left corner. 
In contrast, VRS-HQ leverages the Token-driven Keyframe Selection for more accurate keyframe identification and integrates SAM2 with the temporal token, enriched with both intra-frame spatial and inter-frame temporal relations, resulting in reliable decoding and consistent object tracking.

\textcolor{red}{Fig.}~\ref{result visualization3} presents scenarios requiring general and world knowledge for reasoning.
In the first example (left), VISA segments only two koi carp (Cyprinus carpio) correctly, whereas VRS-HQ identifies nearly all the fish present.  In the second example (right), VISA fails to associate ``dog" with the phrase ``common household pet", indicating limitations in its reasoning capabilities. By contrast, VRS-HQ leverages the integration of temporal tokens to achieve a more nuanced semantic understanding, enabling finer control and interpretation.
% This demonstrates that, even when using the same MLLM, VRS-HQ’s integration and effective use of temporal tokens enable more precise semantic understanding and control compared with VISA.

\section*{F. In-the-wild Visualization Results}
\label{in-the-wild}
\begin{figure*}[t!]
    \centering
    \vspace{-1mm}
    \includegraphics[width=1\linewidth]{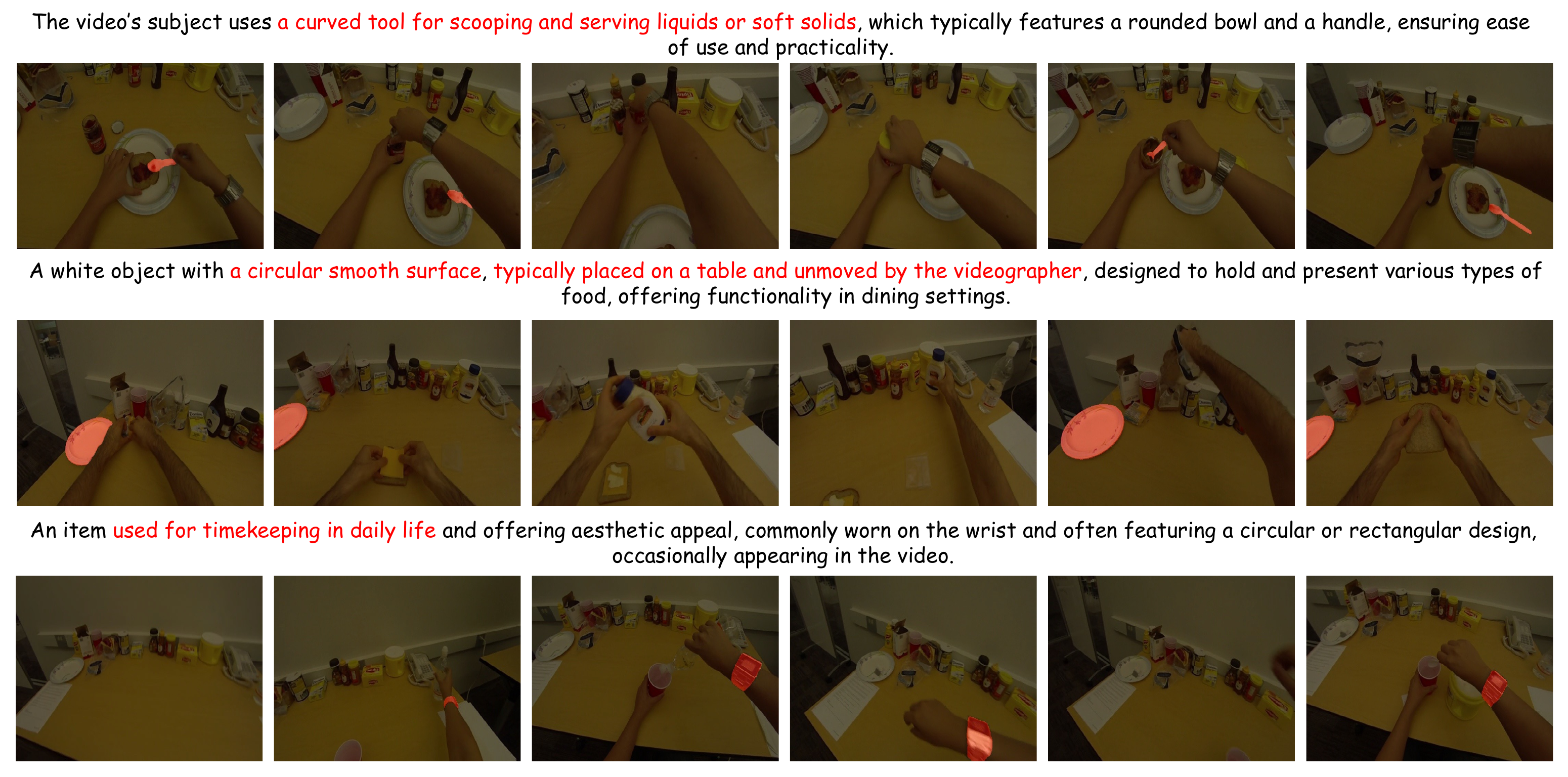}
    \vspace{-5mm}
    \caption{
    Visualization of VRS-HQ utilized in egocentric videos.
    }
    \vspace{-3mm}
    \label{egocentric}
    \vspace{-2mm}
\end{figure*}

\begin{figure*}[t!]
    \centering
    \vspace{-1mm}
    \includegraphics[width=1\linewidth]{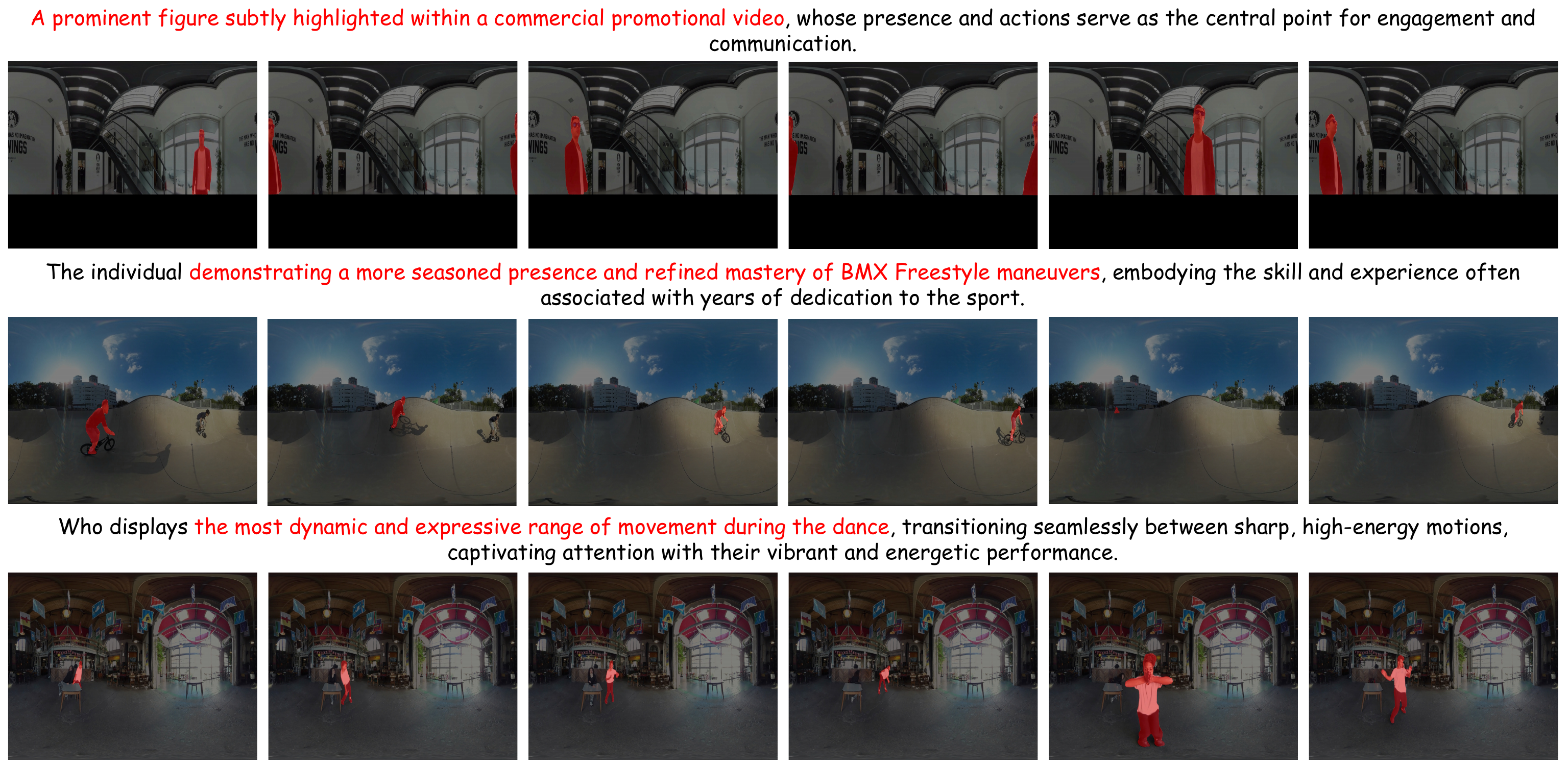}
    \vspace{-5mm}
    \caption{
    Visualization of VRS-HQ applied to 360-degree panoramic videos.
    }
    \vspace{-3mm}
    \label{360panoramic}
    \vspace{-2mm}
\end{figure*}

\textcolor{red}{Fig.}~\ref{egocentric} and \textcolor{red}{Fig.}~\ref{360panoramic} show qualitative results of VRS-HQ on in-the-wild videos. \textcolor{red}{Fig.}~\ref{egocentric} shows results on first-person videos from the GTEA dataset~\cite{fathi2011learning}, using implicit prompts.  Even in cluttered kitchen environments with many similar objects, VRS-HQ demonstrates strong generalization capability.  It is particularly effective at segmenting smaller targets, such as the spoon and watch shown in the first and third rows, respectively, maintaining robust performance in these challenging scenarios. 
\textcolor{red}{Fig.}~\ref{360panoramic} shows results on 360-degree panoramic videos from the PanoVOS dataset~\cite{yan2025panovos}, using more intricate prompts.  Notably, VRS-HQ successfully segments individuals even when they are split across the distorted edges of the video (first row), without any task-specific optimizations.  Furthermore, it maintains effective tracking performance when the primary subjects within the video are moving dynamically (last two rows).

% \clearpage
\bibliographystyle{ieeenat_fullname}
    \bibliography{main}

\end{document}